\documentclass[journal]{IEEEtran}
\usepackage{cite}
\usepackage{url}
\usepackage{hyperref}
\usepackage{array}
\usepackage{graphicx}

\DeclareGraphicsExtensions{.png, .jpg, .pdf, .svg}
\usepackage{amsmath}
\usepackage{multirow}
\usepackage{url}
\usepackage{siunitx}
\usepackage[capitalize]{cleveref}
\usepackage{tikz}
\usepackage{tabularx}
\DeclareUnicodeCharacter{3B2}{$\beta$}
\usepackage{caption}
\begin{document}

\title{Deep Learning Methods for Fingerprint-Based Indoor Positioning: A Review}

\author{Fahad~Alhomayani and Mohammad~H.~Mahoor%
\thanks{Fahad Alhomayani and Dr. Mohammad H. Mahoor are with the Department
of Electrical and Computer Engineering, University of Denver, Denver,
CO, 80208 USA e-mail: \{fahad.al-homayani, mmahoor\}@du.edu.}}%

\markboth{Journal of Location Based Services}{}

\maketitle

\begin{abstract}
Outdoor positioning systems based on the Global Navigation Satellite System have several shortcomings that have deemed their use for indoor positioning impractical. Location fingerprinting, which utilizes machine learning, has emerged as a viable method and solution for indoor positioning due to its simple concept and accurate performance. In the past, shallow learning algorithms were traditionally used in location fingerprinting. Recently, the research community started utilizing deep learning methods for fingerprinting after witnessing the great success and superiority these methods have over traditional/shallow machine learning algorithms. This paper provides a comprehensive review of deep learning methods in indoor positioning. First, the advantages and disadvantages of various fingerprint types for indoor positioning are discussed. The solutions proposed in the literature are then analyzed, categorized, and compared against various performance evaluation metrics. Since data is key in fingerprinting, a detailed review of publicly available indoor positioning datasets is presented. While incorporating deep learning into fingerprinting has resulted in significant improvements, doing so, has also introduced new challenges. These challenges along with the common implementation pitfalls are discussed. Finally, the paper is concluded with some remarks as well as future research trends.
\end{abstract}


\IEEEpeerreviewmaketitle

\section{Introduction}
\IEEEPARstart{O}{ver} the past two decades, the limitations satellite-based outdoor positioning systems (e.g., GPS, Galileo, GLONASS) have for indoor use \cite{8409950} led researchers to propose a wide variety of indoor positioning systems. Indoor positioning or indoor localization is the process of determining one's indoor location with respect to a predefined frame of reference. Indoor navigation relies on positioning updates to reach a target location from the current location. All indoor positioning systems are designed to provide location information. Some go a step further to provide navigation capabilities.

While the notion of location is broad, location information can generally be presented in one of four ways: physically, absolutely, relatively, and symbolically \cite{hightower2001location, liu2007survey}. Physical location is obtained with respect to a global reference frame (e.g., latitude and longitude in the geographic coordinate system). Absolute location is expressed with respect to a local reference frame and the resolution of the frame depends on grid size. Relative location expresses the user's proximity to known landmarks in the environment. Symbolic location expresses location in a natural-language way, thus, providing abstract information of where the user is (e.g., in the living room, in the kitchen, etc.).

A common theme in early indoor positioning systems is an infrastructure-based nature. In other words, early systems provide positioning by relying on specialized equipment that has to be deployed throughout the environment and carried by users. Such equipment include ultrasonic transmitters, infrared badges, and Radio Frequency IDentification (RFID) tags \cite{hightower2001location,liu2007survey}. In contrast, the most recent systems are either infrastructure-free or take advantage of the already deployed infrastructure (e.g., WiFi Access Points (APs)). These systems rely on the various sensors and modules found in users' smartphones to provide indoor positioning \cite{subbu2014analysis, davidson2017survey}. Infrastructure-free positioning systems do not necessitate deployed hardware in the environment to operate. Examples of such systems include magnetic field-based systems and camera-based systems (if artificial markers are not required for positioning).

\begin{figure}[!t]
\centering
\includegraphics[width=0.55\columnwidth]{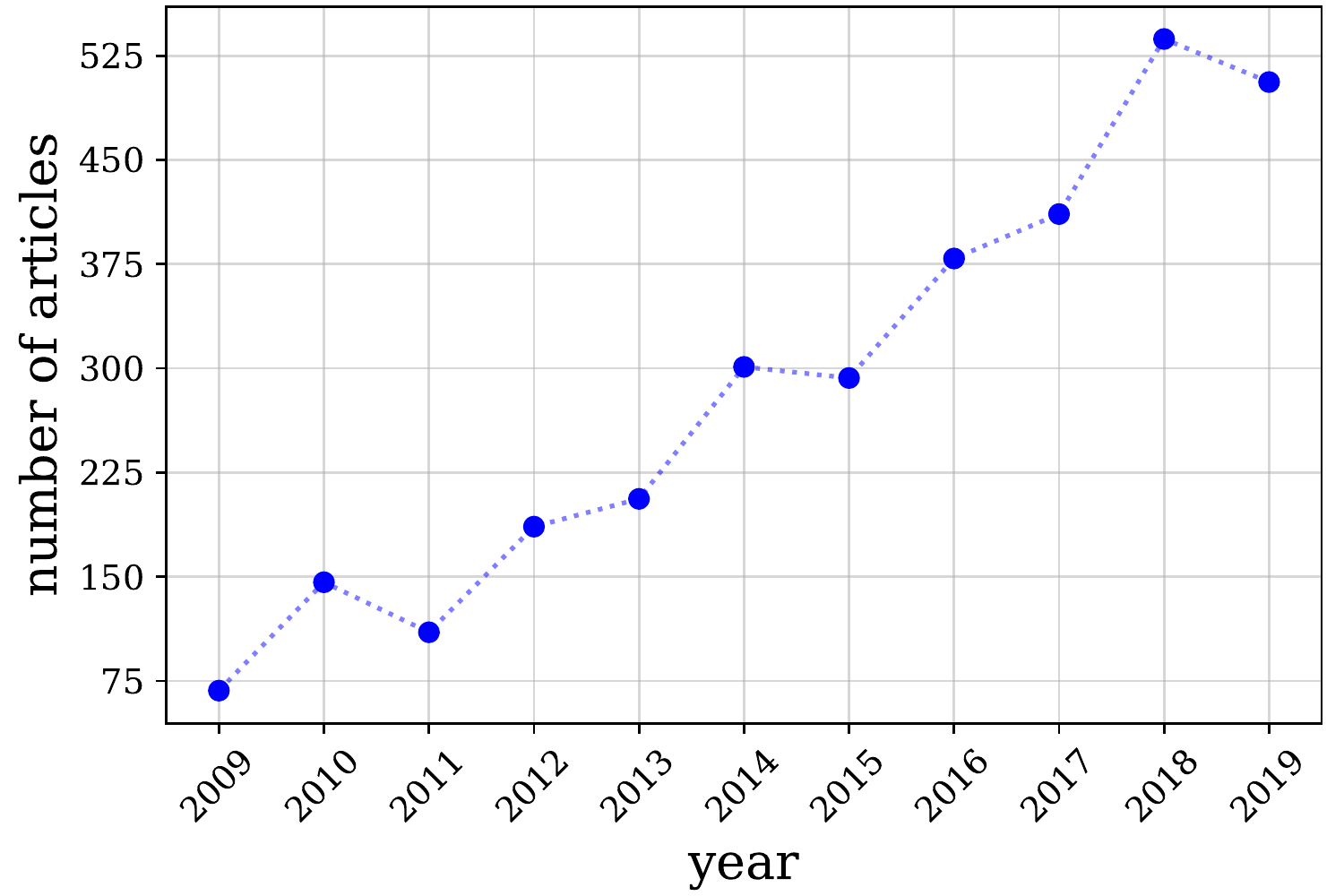}
\caption{The number of published articles in IEEE Xplore by year (from \num{2009} to \num{2019}) where authors used ``indoor positioning'', ``indoor localization'', or ``indoor navigation'' as a keyword.}
\label{article_vs_year}
\end{figure}

Designing an indoor positioning system has remained a challenging task since indoor environments are very complex and are often characterized by non-line-of-sight (NLoS) settings, moving people and furniture, walls of different densities, and the presence of different indoor appliances that alter indoor signal propagation. Nevertheless, the demand for more complete solutions is higher than ever before. This demand is fueled by a multitude of potential applications and services enabled by indoor positioning. Indoor positioning is a key enabling technology for many domains including indoor location-based services (ILBS) \cite{HuangandGeorg}, the internet of things (IoT) \cite{macagnano2014indoor}, ambient assisted living (AAL) \cite{rashidi2013survey}, indoor emergency responders navigation \cite{ferreira2017localization}, and occupancy detection for the energy-efficient control of buildings \cite{CHEN2018260}. Attempting to satisfy the demand, researchers are forced to compromise between different design criteria (e.g., accuracy, precision, privacy, scalability, complexity, cost, etc.\cite{liu2007survey}). To date, no universally agreed upon solution has emerged to solve the indoor positioning problem. Because of this, indoor positioning research is vibrant. Researchers share their work in dedicated conferences such as, the International Conference on Indoor Positioning and Indoor Navigation (IPIN); the International Conference on Ubiquitous Positioning, Indoor Navigation and Location-Based Services (UPINLBS); and the Workshop on Positioning, Navigation and Communication (WPNC). As seen in Fig. \ref{article_vs_year}, the body of literature published in these conferences' proceedings, as well as at other venues and in other journals, continues to grow each year.

This paper provides a comprehensive review of deep learning methods for fingerprint-based indoor positioning with the objective of covering the developments of this area of research from its inception to its current state and beyond. Wherever appropriate, topics are presented in a chronological context and special emphasis is given to clear pioneers and major milestones.

\subsection{The Fingerprinting Approach to Indoor Positioning}

Various approaches for indoor positioning have been proposed over the years. The main methods introduced include \textit{angulation}, \textit{lateration}, \textit{proximity detection}, \textit{pedestrian dead reckoning}, and \textit{location fingerprinting}. Amongst these, the latter has recently received significant attention as a straightforward, inexpensive, and accurate approach for indoor positioning. Location fingerprinting, also referred to as \textit{scene analysis}, or fingerprinting, employs low-power sensors that are integrated into smartphones and exploits existing infrastructure, such as WiFi APs, to achieve high positioning accuracy even in NLoS settings. The location of these APs is not a prerequisite for positioning, which eliminates the need to model complex indoor signal propagation \cite{MokandG}. Moreover, fingerprinting systems are immune to accumulated positioning errors caused by Inertial Measurement Unit (IMU) drifts \cite{JackermeierandBernd}. 

The concept of fingerprinting is identifying indoor spatial locations based on location-dependent measurable features (location fingerprints). There are different types of fingerprints such as \textit{radio frequency} fingerprints \cite{RADAR}, \textit{magnetic field} fingerprints \cite{886523}, \textit{image} fingerprints \cite{729529}, and \textit{hybrid} fingerprints \cite{Azizyan_2009}. Radio frequency fingerprints, particularly WiFi fingerprints, are, undoubtedly, the most used fingerprints.

\begin{figure*}[!h]
\centering
\includegraphics[width=1.2\columnwidth]{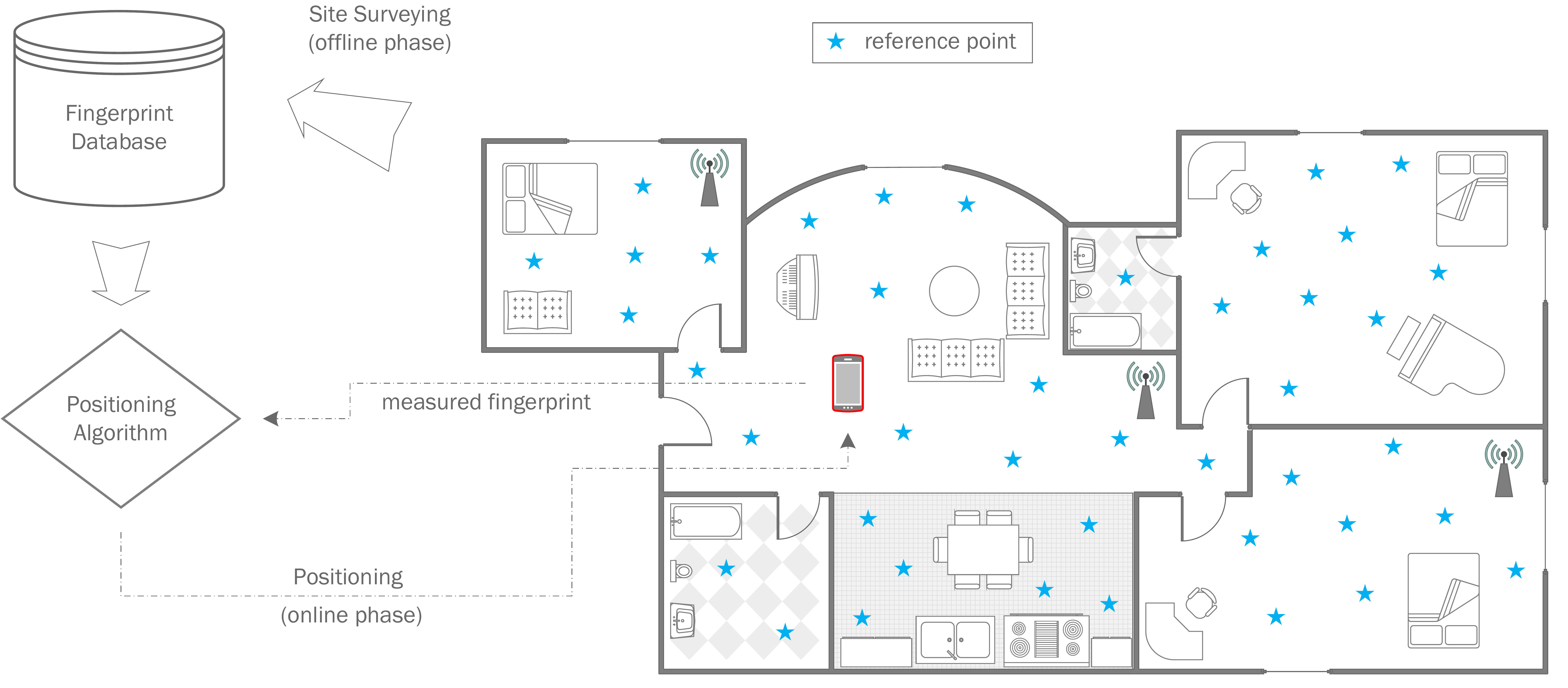}
\caption{A graphical representation of the fingerprinting approach to indoor positioning.}
\label{fingerprint_process}
\end{figure*}

From an implementation perspective, the fingerprinting approach to indoor positioning is a two-phase process that consists of an \textit{offline phase} and an \textit{online phase}.  During the offline phase, \textit{site surveying}, in which the fingerprints of the area of interest are sampled at predefined \textit{reference points} (RPs), is performed. The fingerprints are sampled using smartphone sensors. For example, the WiFi module and the magnetometer are used to collect received signal strength (RSS) and magnetic field fingerprints, respectively. The sampled fingerprints, along with their corresponding coordinates, are stored in a database. The data is then used to train a machine learning algorithm to learn a function that best maps the sampled fingerprints to their correct coordinates. The learned function is then used during the online phase to infer a user's coordinates given the measured fingerprints at the user's location. The process of fingerprinting is visually depicted in Fig. \ref{fingerprint_process}.

The main source of error in fingerprinting systems is due to \textit{location ambiguity}. Location ambiguity refers to the problem of different RPs exhibiting similar fingerprints \cite{luo2017indoor}. Local ambiguity occurs when adjacent RPs have similar fingerprints, while global ambiguity occurs when distant RPs have similar fingerprints. As discussed later, different fingerprint types may suffer from one ambiguity more than the other. For example, WiFi fingerprints are generally immune to global ambiguity but prone to local ambiguity, while the contrary is true for magnetic field fingerprints.

Based on the number of samples needed for online positioning, a given system can be classified as either \textit{one-shot} or \textit{multi-shot} \cite{7103024}. In a one-shot system, a location is estimated using only a single fingerprint sample; while in a multi-shot system, two or more samples (i.e., consecutive measurements) are required to refine the positioning estimate. Due to the time spent obtaining the additional samples and the pre/post-processing involved, multi-shot systems are generally slower but more accurate than one-shot systems.

Shallow learning algorithms such as $k$-Nearest Neighbor ($k$NN), Naïve Bayes, and Decision Trees have traditionally been utilized for location fingerprinting \cite{PereiraandChristianT,MirowskiandPhilip,JTandZT,ZhouandAndreasW}. The research community is rapidly shifting towards deep learning-based fingerprinting after witnessing the tremendous success that deep learning methods have achieved in a multitude of research fields and applications. 

\subsection{Why Deep Learning for Fingerprinting}
Listed below are some powerful deep learning algorithms properties and their positive implications on location fingerprinting:
\begin{enumerate}
\item Deep learning techniques often provide an end-to-end solution where the task of feature extraction is automatically performed and implicitly embedded in the architecture, avoiding the need for hand-engineered features, a time-consuming and knowledge-demanding process. This property is particularly crucial when dealing with high-dimensional and not-easily extractable features that are required for radio frequency and image fingerprinting.
\item Deep learning is well-known for effectively and efficiently processing massive amounts of raw data, a task otherwise difficult, if not impossible. In fact, the predictive performance of deep learning algorithms enhances with increased training samples. Consequently, there is no limit to the amount of fingerprint data used for training.
\item The parametric nature of deep learning, where computational complexity does not depend on dataset size and the ability to parallelize computation using Graphical Processing Units (GPUs) results in infinitesimal inference latency (in the orders of milliseconds or less), makes deep learning algorithms ideal for real-time positioning applications. However, this often comes at the expense of a prolonged training phase.
\item Deep learning is the method of choice for classification/regression problems in which the nature of boundaries describing the features in input space is highly complex and nonlinear. This is the case in fingerprinting where the overarching goal is to distinguish between spatial locations that are, in many cases, separated by a few centimeters or less.  
\item Deep learning is well-suited for transfer learning which involves transferring knowledge from pre-trained networks to minimize data collection and training efforts. Therefore, a fingerprinting system can be realized with minimal cost. In this regard, unsupervised and semi-supervised deep learning methods have also proven successful when the fingerprint data is scarce or unlabeled.
\end{enumerate}
 
\section{Review Scope, Related Work, and Contributions}
\subsection{Review Scope}
While deep learning started to gain momentum in \num{2006}, after Hinton \textit{et al}.\cite{hinton2006fast} made a historic breakthrough in training deep architectures, the intersection between deep learning and fingerprinting didn't take place until almost a decade later. The exponentially increasing body of literature ever since motivated the composition of this review. The aim of this review is to provide researchers and practitioners with the current solutions and how they compare, potentials, and challenges of this ever-expanding area of research. To this end, over \num{40} research papers, published between \num{2015} and \num{2019}, where deep learning methods were leveraged for fingerprinting, were identified. These papers were selected based on scientific quality, originality, and significance. Scientific quality is ensured by reviewing a paper only if it was published in a peer-reviewed, reputable journal or conference proceedings. The originality criterion ensures that priority is given to papers studying topics that have received little to no attention. In this sense, if two identified papers discuss very similar topics, then a higher priority for inclusion is given to the paper that was published first. Significance is assessed by means of citation counts. Specifically, if a paper has not received any non-self citations within \num{18} months of publication, it receives lower priority for inclusion. The scope of the review is specific in nature since this paper does not review fingerprinting based on shallow learning nor does it review areas where deep learning is used with other positioning methods. The deep learning methods covered in this paper are Autoencoders (AEs), Convolutional Neural Networks (CNNs), Deep Belief Networks (DBNs), Fully Connected (FC) Networks, Generative Adversarial Networks (GANs), and Recurrent Neural Networks (RNNs). Providing a discussion on these methods is beyond the scope of this paper. Readers looking for details about deep learning may refer to \cite{goodfellow2016deep}. The acronyms and abbreviations used throughout the paper are listed in Table \ref{Acronyms_and_Abbreviations}.

\begin{tiny}
\begin{table*}[!t]
\tiny
\renewcommand{\arraystretch}{1.3}
\caption{acronyms and abbreviations}
\label{Acronyms_and_Abbreviations}
\centering
\begin{tabular}{m{0.2cm}m{2.2cm}}
AAL&Ambient Assisted Living\\
AE&Autoencoder\\
AoA&Angle of Arrival\\
AP&Access Point\\ 
BIM&Building Information Modeling\\
BLE&Bluetooth Low Energy\\
BS&Base Station\\
CDF&Cumulative Distribution Function\\
CGAN&Conditional Generative Adversarial Network\\
CIR&Channel Impulse Response\\
CNN&Convolutional Neural Network\\
CPU&Central Processing Unit\\
CSI&Channel State Information\\
DAE&Denoising Autoencoder\\
\end{tabular}
\quad
\begin{tabular}{m{0.5cm}m{2.2cm}}
\si{dBm}&decibel-milliwatts\\
DBN&Deep Belief Network\\
DRL&Deep Reinforcement Learning\\
FC&Fully Connected\\
FLOP&Floating-Point Operation\\
GAN&Generative Adversarial Network\\
GLONASS & GLObal NAvigation Satellite System\\
GPS&Global Positioning System\\
GPU&Graphical Processing Unit\\
GRU&Gated Recurrent Unit\\
LAR&Locomotion Activity Recognition\\
HMM&Hidden Markov Model\\
ILBS&Indoor Location-Based Services\\
\end{tabular}
\quad
\begin{tabular}{m{0.2cm}m{2.2cm}}
IMU&Inertial Measurement Unit\\
IoT&Internet of Things\\
IPIN&The International Conference on Indoor Positioning and Indoor Navigation \\
$k$NN&$k$-Nearest Neighbor\\
LED&Light Emitting Diode\\
LoS&Line-of-Sight\\
LSTM&Long Short-Term Memory\\
LTS&Localization and Tracking System\\
MAE&Mean Absolute Error\\
MIMO&Multiple-Input Multiple-Output\\
MSE&Mean Squared Error\\
\si{\micro\tesla}&microtesla\\
\end{tabular}
\quad
\begin{tabular}{m{0.2cm}m{2.2cm}}
M2M&Machine-to-Machine\\
NLoS&Non-Line-of-Sight\\
OFDM&Orthogonal Frequency-Division Multiplexing\\
ONNX&Open Neural Network Exchange\\
RFID&Radio Frequency IDentification\\
RGB&Red-Green-Blue\\
RMSE&Root Mean Squared Error\\
RNN&Recurrent Neural Network\\
RP&Reference Point\\
RSS&Received Signal Strength\\
SAE&Stacked Autoencoder\\
SDAE&Stacked Denoising Autoencoder\\
\end{tabular}
\quad
\begin{tabular}{m{0.4cm}m{2.4cm}}
SfM&Structure-from-Motion\\
SIFT&Scale-Invariant Feature Transform\\
SURF&Speeded Up Robust Features\\
SVM&Support Vector Machine\\
TTFF&Time-To-First-Fix\\
URLLC&Ultra-Reliable and Low-Latency Communication\\
UUID&Universally Unique Identifier\\
UWB&Ultra-Wide Band\\
VAE&Variational Autoencoder\\
VLC&Visible Light Communication\\
W$k$NN&Weighted $k$-Nearest Neighbor\\
WNIC&Wireless Network Interface Card\\
WSN&Wireless Sensor Network\\
\end{tabular}
\end{table*}
\end{tiny}

\subsection{Related Work}
Since the field of indoor positioning is not novel, over the years several articles have been published that generally review the field \cite{liu2007survey, CurranandEoghan, TangandJohn, yassin2016recent, 8692423} or review it from different angles such as the positioning approach used \cite{harle2013survey, he2016wi, guvenc2009survey}, the underlying technology utilized \cite{he2016wi,luo2017indoor,pasku2017magnetic}, or the application domain tackled \cite{ferreira2017localization, alvarez2013evaluation}. While all these works are remarkable, none of them discussed deep learning-based indoor positioning.

From a deep learning standpoint, Mohammadi \textit{et al}. \cite{mohammadi2018deep} and Zhang \textit{et al}. \cite{zhang2018deep} recently reviewed deep learning approaches and use cases in the context of IoT big data and streaming analytics, and mobile and wireless networks, respectively. Both works marginally introduced deep learning-based indoor positioning. However, an in-depth review where solutions are analyzed, categorized, and compared was not provided.

To the extent of our knowledge, this is the first study dedicated to reviewing the recent adoption of deep learning methods in fingerprint-based indoor positioning.

\subsection{Contributions and Paper Organization}
This article’s contributions are in line with its general organization:
\begin{itemize}
\item Section \hyperref[sec2]{III} overviews various fingerprint types and discusses their advantages and disadvantages for indoor positioning.
\item Since data is at the core of every fingerprinting system, whether based on deep or shallow learning, Section \hyperref[sec3]{IV} provides an elaborate review of indoor positioning datasets that are currently publicly available. Using different variables, datasets are compared to help researchers and practitioners choose the dataset that best fits their implementation goals.
\item Section \hyperref[sec4]{V} introduces a performance evaluation framework for deep learning-based fingerprinting systems. The framework consists of five metrics that can be used to evaluate the quality of a given system from different aspects.
\item Section \hyperref[sec5]{VI} thoroughly investigates current deep learning-based fingerprinting solutions proposed in the literature. For the convenience of analysis, these solutions are categorized based on the fingerprint type they employ and subcategorized based on the deep learning model they exploit. Additionally, all solutions are evaluated and compared using the evaluation framework described in Section \hyperref[sec4]{V}.
\item Having reviewed the literature, Section \hyperref[sec6]{VII} identifies common pitfalls to avoid when designing a deep learning-based fingerprinting system and highlights the implementation challenges that have yet to be addressed.
\item Section \hyperref[sec7]{VIII} suggests future research directions and concludes the review.
\end{itemize}

\section{Indoor Fingerprint Types}
\label{sec2}
This section provides an overview of different fingerprint types that are used for indoor positioning. For  each  fingerprint  type, its advantages and disadvantages for indoor positioning are discussed first, followed by a brief account of the first documented time of using it for indoor positioning. The fingerprint types include Radio Frequency (WiFi, BLE, and Cellular), Magnetic Field and IMU, Image, Hybrid, and Miscellaneous (Ultra-Wide Band (UWB), Visible Light, RFID, and Acoustic).

\subsection{Radio Frequency Fingerprints}

\subsection*{1) WiFi Fingerprints}
The family of IEEE \num{802.11} Wireless Local Area Network (WLAN) standards, commonly known as WiFi, operate in two unlicensed bands: the \SI{2.4}{\giga\hertz} and \SI{5}{\giga\hertz} bands. WiFi was designed to provide high-speed wireless networking and Internet connectivity; thus, it is optimized for communication rather than localization. Nevertheless, using WiFi for localization is a natural choice because of its widespread adoption in user devices and the ubiquity of WiFi APs. Moreover, no additional infrastructure is required to realize localization, making WiFi fingerprinting a cost-effective solution.

WiFi fingerprints are formed by extracting RSS values from all visible APs in an environment. Thus, one drawback of WiFi fingerprinting is the time it takes to complete a scanning cycle. Depending on hardware/software limitations, this process can take several seconds \cite{KAEMARUNGSI2012292}. This becomes problematic when the user is moving.  Movement may lead to smearing the fingerprint across space \cite{7103024}. Another drawback of using WiFi fingerprints is associated signal interference. Many indoor appliances such as microwave ovens, cordless phones, and wireless baby monitors operate in the same bands as WiFi. This often leads to high variability in RSS measurements, even when recorded at the same location \cite{KAEMARUNGSI2012292,6761255,BaiandSuqin}.

In \num{2000}, Microsoft Research proposed \textit{RADAR} \cite{RADAR}, a system  widely known as the first WiFi fingerprinting system. The system collects RSS measurements at the AP side instead of the user side; thus, it is a tracking system. The $k$NN algorithm, with a Euclidean distance similarity metric, is used to compute a user’s position. RADAR designers demonstrated that a user’s orientation, the value of $k$, and the number of samples in the offline and online phases affect localization accuracy. The superiority of fingerprinting over lateration was also demonstrated. Fingerprinting achieved a median localization error of \SI{2.94}{\meter} compared to \SI{4.3}{\meter} achieved by lateration. Later, a Viterbi-like algorithm was proposed to enhance the system’s tracking ability \cite{bahl2000enhancements}. The median error was reduced to \SI{2.37}{\meter}.

\begin{figure}[!b]
\centering
\includegraphics[width=0.45\columnwidth]{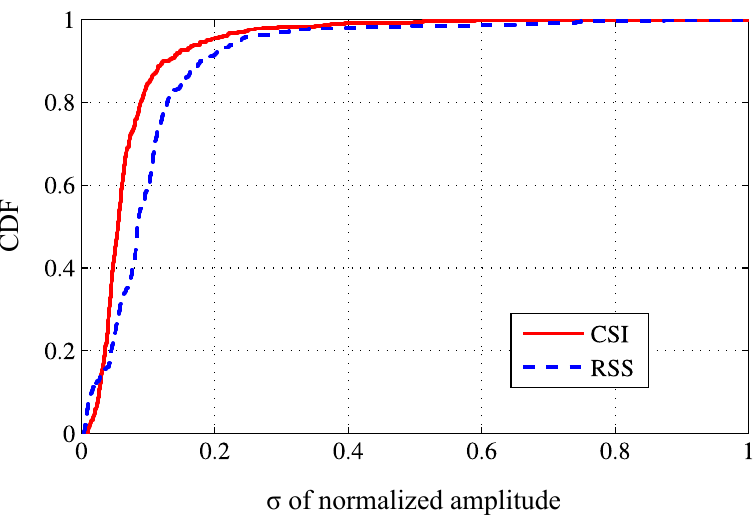}
\caption{CDF of the standard deviations of CSI and RSS amplitudes for \num{150} locations using \num{50} measurements at each location. Figure reproduced from \cite{DeepFi2}.}
\label{CSI}
\end{figure}

Currently, there is a trend in exploiting richer information enabled by orthogonal frequency-division multiplexing (OFDM) through Channel State Information (CSI). CSI includes the amplitude and phase of each subcarrier from each antenna. CSI is a function of the combined effect of multipath, shadowing, power decay, and fading on a signal propagating from a transmitter to receiver. Since many subcarriers are available for each antenna, positioning using a single AP is feasible \cite{PinLoc,FIFS}. Moreover, CSI values have proven to be more stable than RSS values as demonstrated in Fig. \ref{CSI}. However, the main drawback of using CSI for fingerprinting is that most WNICs do not provide means for conveniently extracting CSI values. Impractical solutions, such as hacking into device drivers, are commonly followed for data collection. At the time of writing this paper, no implementation that uses a smartphone to collect CSI data exists. 

\subsection*{2) BLE Fingerprints}
Bluetooth Low Energy (BLE), also known as Bluetooth Smart or Bluetooth \num{4.0}, is a popular wireless technology for low-power, machine-to-machine (M2M) communication. It has \num{40}, \SI{2}{\mega\hertz} wide channels that operate in the same \SI{2.4}{\giga\hertz} radio band as WiFi \cite{bluetooth}. Since the Bluetooth Special Interest Group introduced it in \num{2010}, it has received widespread adoption with over \num{800} million BLE-enabled devices shipped in \num{2019} alone \cite{bluetooth_market}. One of the main driving forces behind its popularity are BLE beacons. BLE beacons are small, inexpensive, and portable (battery-powered) transmitters that are used in a multitude of applications, including indoor positioning. Some beacons allow for the adjustment of transmission parameters such as transmission frequency, power, and bit rate. Beacons use three widely spaced channels to broadcast advertising messages that contain the beacon’s Universally Unique Identifier (UUID) and its transmission power in decibel-milliwatts (\si{dBm}). These messages are used by proximity-based positioning systems to provide positioning and navigation services. \cite{8242361}. Two widely used industry protocols for BLE include Apple's iBeacon and Google’s Eddystone.

Regarding fingerprinting, Faragher and Harle \cite{7103024} investigated the feasibility of using BLE fingerprints for fine-grained indoor positioning. They conducted extensive experiments from which they reported several findings. First, the power draw on smartphones is much lower for BLE than WiFi. Second, BLE has a much higher scan rate than WiFi which makes BLE more suitable for user navigation and tracking applications. Third, if enough BLE beacons are strategically deployed in an environment, then the positioning accuracy could easily surpass that obtained by the existing WiFi infrastructure. However, BLE signals are more vulnerable to channel gain and fast fading than WiFi signals. As a result, BLE measurements fluctuate severely over time. The use of three channels (compared to one in WiFi) exacerbates this problem due to the wide spacing between these channels. Additionally, monitoring the battery level of the deployed BLE beacons to ensure uninterrupted services is still a major challenge \cite{8242361}. Table \ref{WiFi_AP_vs_BLE_beacon} compares some of the technical specifications of a typical WiFi AP and BLE beacon.

\begin{table}[!b]
\tiny
\renewcommand{\arraystretch}{1.3}
\caption{WiFi AP vs. BLE beacon}
\label{WiFi_AP_vs_BLE_beacon}
\centering

\begin{tabular}{|l|c|c|}
\cline{2-3}
\multicolumn{1}{c|}{} & WiFi AP$^{\dag}$ & BLE beacon$^{\ddag}$  \\ \hline
Battery powered & No   & Yes    \\ \hline
Max. power consumption (\si{\watt}) & \num{12.7} & \num{0.01}  \\ \hline
Max. transmit power (\si{dBm}) & \num{20}  & \num{0}     \\ \hline
Max. range (\si{\meter}) & \num{250}  & \num{50}     \\ \hline
Weight (\si{\kilo\gram}) & \num{1.020}  & \num{0.047}     \\ \hline
Cost (\si{\$}) & $\approx$ \num{100.00} & $\approx$ \num{30.00}  \\ \hline
\multicolumn{3}{l}{$^{\dag}${\tiny TP-Link EAP245 AP} $^{\ddag}${\tiny Aruba LS-BT20 beacon}}\\

\end{tabular}
\end{table}

\subsection*{3) Cellular Fingerprints}
The use of cellular-based indoor positioning has primarily been motivated by the E-\num{911} regulation imposed by the U.S. Federal Communications Commission (FCC) \cite{8226757}. The most recent regulation mandates require cellular network operators to provide emergency call positioning within a \SI{50}{\meter} horizontal accuracy \cite{e911_requirements_2015} and \SI{3}{\meter} vertical accuracy \cite{e911_requirements_2019}. Due to the lack of access to proprietary cellular data, such as time and angle measurements, most academic solutions to cellular indoor positioning are either fingerprinting- or triangulation-based \cite{8409950}. 

\begin{figure}[!b]
\centering
\includegraphics[width=0.6\columnwidth]{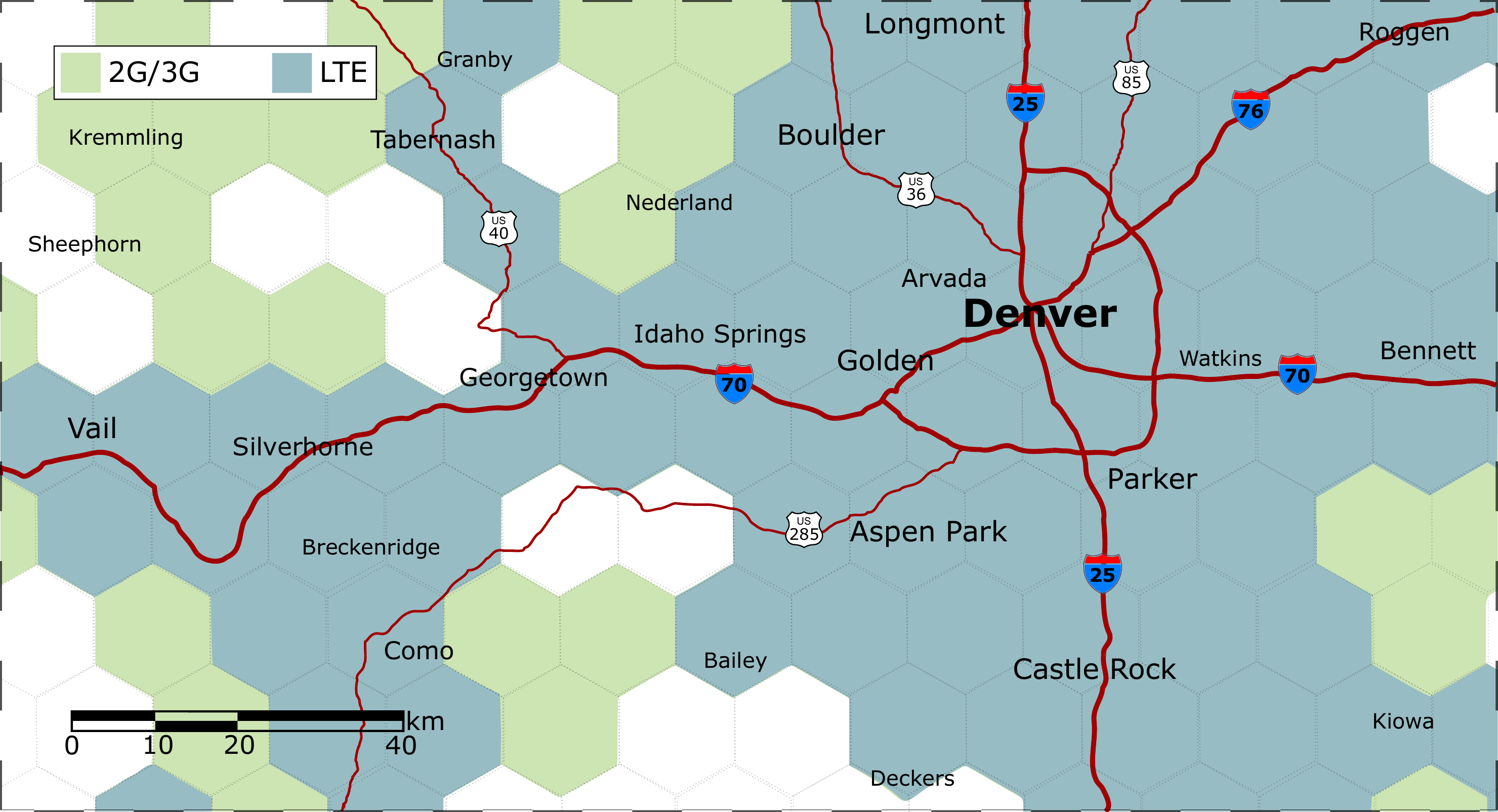}
\caption{Macro-cell layout of a cellular network provider in the U.S. for a selected area inside the state of Colorado. Data obtained from \cite{coveragemap}.}
\label{macrocell}
\end{figure}

From a fingerprinting perspective, cellular-based fingerprinting has several advantages over WiFi/BLE fingerprinting. First, unlike WiFi and BLE, cellular signals operate in licensed bands which means they are less prone to interference. Second, not every cellphone necessarily supports WiFi/BLE; however, every cellphone, by definition, comes equipped with a cellular modem. Third, the typical coverage of cellular base stations (BSs) ranges from hundreds of meters to tens of kilometers which is orders of magnitude greater than WiFi APs/BLE beacons. Fourth, there is no deployment cost associated with using cellular signals for fingerprinting since BSs are deployed and maintained outside the localization environment. Nonetheless, cellular fingerprinting has its drawbacks: First, cellular signals are not designed to penetrate deep inside buildings, often resulting in blind spots due to the shadowing effect. Second, BSs are often deployed on macro-cell layouts (Fig. \ref{macrocell}) in which the overlap between the coverage area of neighboring BSs is kept to a minimum \cite{8226757}, resulting in few fingerprints for any given area. Third, standard-compliant modems can only report the RSS measurements from up to seven BSs \cite{6062428}, limiting the number of measured fingerprints to seven at any given time.

Historically, the first to exploit cellular RSS fingerprints for indoor positioning was Otsason \textit{et al}. in \num{2005} \cite{10.1007/11551201_9}. They used a special modem that provided RSS measurements from up to \num{35} \num{2}G BSs. Experimental results conducted in three buildings demonstrated a median positioning error ranging from \SI{2.48}{\meter} to \SI{5.44}{\meter} using the $k$NN algorithm.

\subsection{Magnetic Field and IMU Fingerprints}
\begin{figure}[!t]
\centering
\includegraphics[width=0.65\columnwidth]{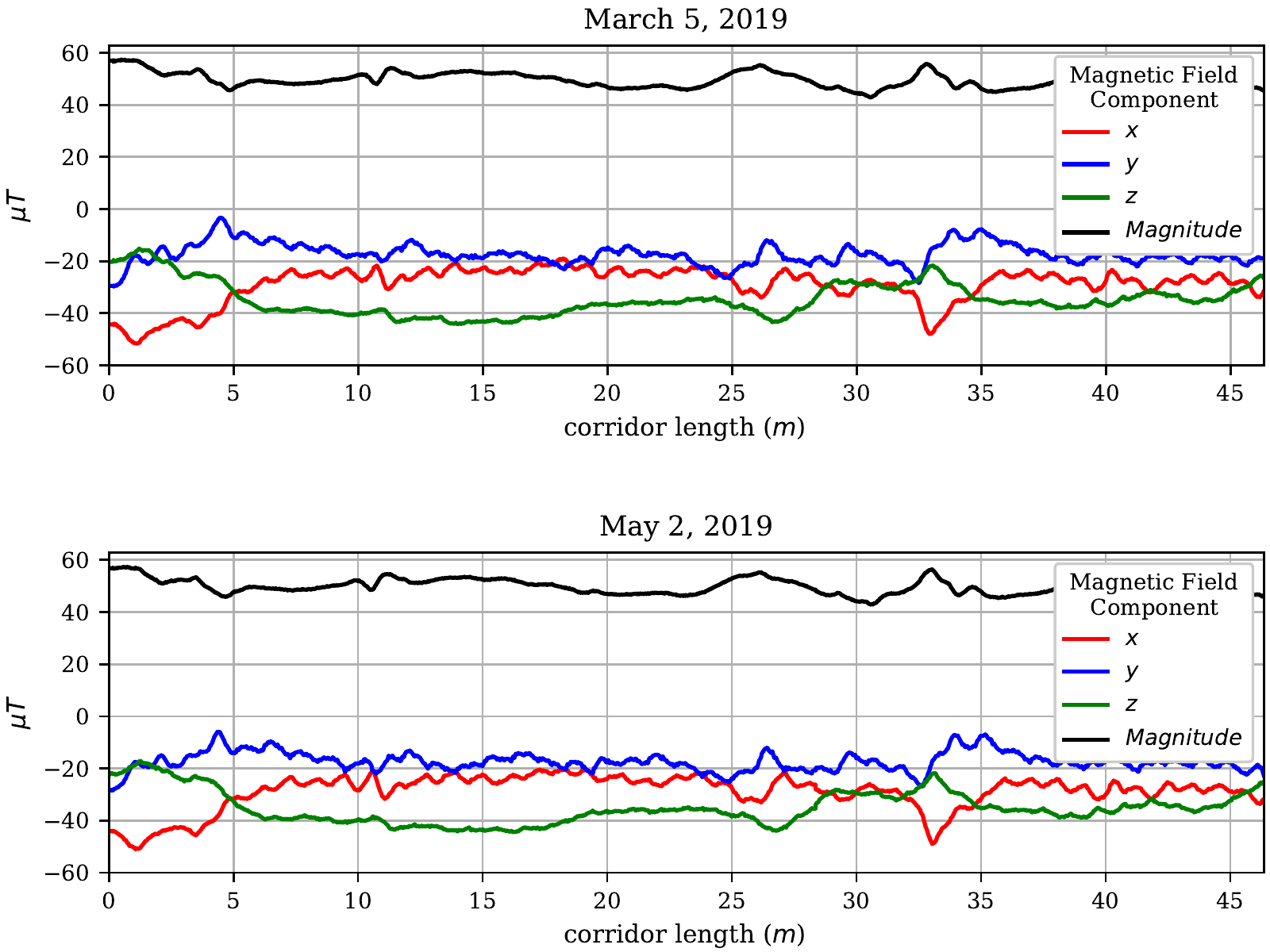}
\caption{Two measurements taken two months apart of the magnetic field strength along a \SI{46}{\meter} long corridor.}
\label{mag_xyz}
\end{figure}

The complex distortions of Earth's magnetic field, caused by steel structures and reinforced concrete, form unique spatial signatures that can be used to construct magnetic maps of indoor environments. These signatures have been experimentally proven to be very stable over long periods \cite{gozick2011magnetic}. They have also been proven to vary significantly across space (in the orders of a few centimeters or less) \cite{li2012feasible}. This property of temporal stability and spatial instability, as depicted in Fig. \ref{mag_xyz}, provides the basis for using the distortions as location fingerprints.

\begin{figure}[!b]
\centering
\includegraphics[width=0.25\columnwidth]{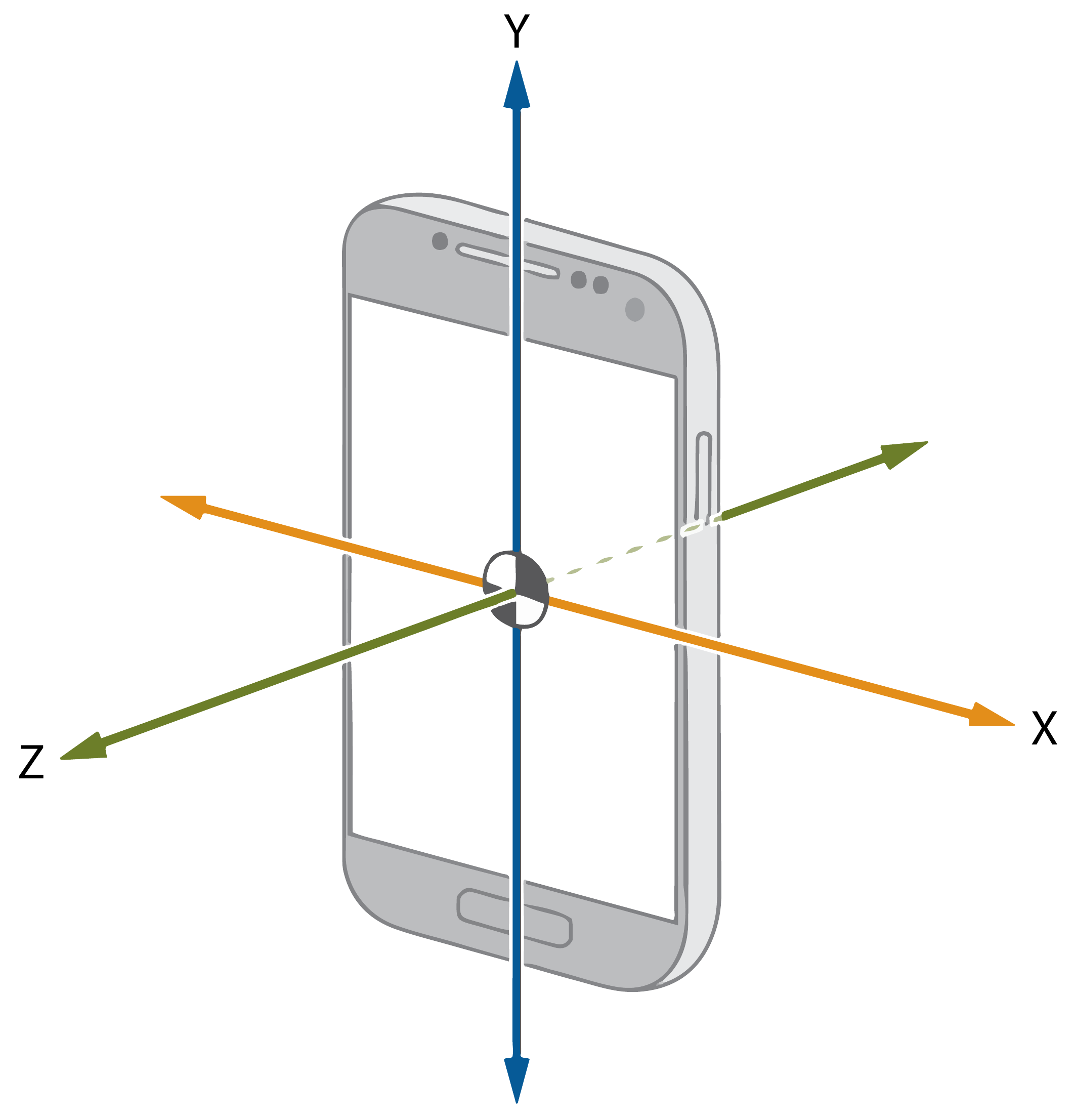}
\caption{Illustration of the X, Y, and Z axes relative to a typical smartphone. Figure adapted from \cite{ReferenceFrame}.}
\label{phone}
\end{figure}

Magnetic field fingerprints are omnipresent and do not require the deployment of special infrastructure, such as APs in the case of RSS fingerprinting, to be realized. Moreover, a smartphone's magnetometer, which measures fingerprints in microtesla (\si{\micro\tesla}), consumes far less energy than its WiFi or Bluetooth modules \cite{subbu2014analysis}. As a result, magnetic field fingerprinting has attracted researchers since it appears to be a promising alternative for indoor positioning. However, most smart devices come equipped with triaxial magnetometers, meaning that the resultant fingerprints only have three features. These features are orientation-dependent because they are measured with respect to the device's reference frame (Fig. \ref{phone}). Consequently, the features are further reduced to two if no restrictions are posed on a smartphone's orientation during the online phase. An orientation independent measure is the magnitude of the magnetic field. However, the magnitude is a single component and using it as a fingerprint can lead to global ambiguity. Another drawback of a magnetic field fingerprint is the vulnerability to magnetic interference caused by objects such as elevators and vending machines.

Among the first to realize that an electronic compass’ incorrect heading information can be used as a signature for indoor localization was Suksakulchai \textit{et al}. in \num{2000} \cite{886523}. They mounted an electronic compass on top of a service robot ``HelpMate'' and collected the heading information as the robot traversed a corridor. The next time the robot traversed the corridor, it matched its measured heading information with the pre-collected information; if a match was found, the robot could determine its position. In \num{2011}, Gozick \textit{et al}. \cite{gozick2011magnetic} used mobile phones’ built-in magnetometers  to build magnetic maps of corridors inside buildings. These maps were constructed with the phones' $y$-axes parallel to the north and prior knowledge of the corridors’ steel pillars locations. The authors used the magnitude of the magnetic field as a feature to differentiate between the different pillars (magnetic landmarks). They showed that the magnetic signatures collected by different mobile phones with different sampling rates have the same pattern.

\subsection{Image Fingerprints}
Using images for indoor localization is viable because most smart devices are armed with cameras. Like magnetic- and cellular-based localization, image-based localization does not depend on infrastructure for operation. Nonetheless, in some scenarios, cameras may not be allowed indoors due to privacy and security concerns \cite{7759302}. Furthermore, image fingerprints are the largest in terms of memory footprint and number of features. For example, compare an image fingerprint captured by an iPhone \num{7}, a fingerprint with \num{12} million features and a memory footprint of \num{6} MB (stored as a \texttt{.jpg} file), to a WiFi fingerprint with \num{127} features and a memory footprint of \num{4} KB (stored as a \texttt{.txt} file). Therefore, to reduce the number of features for training, image-based localization systems often re-size images to a lower resolution and use cropping to select only the region of interest. Additionally, image compression techniques should be considered when relying on a remote server for positioning or when the available bandwidth for transmission is limited \cite{5888650}.

\begin{figure}[!b]
\centering
\includegraphics[width=0.8\columnwidth]{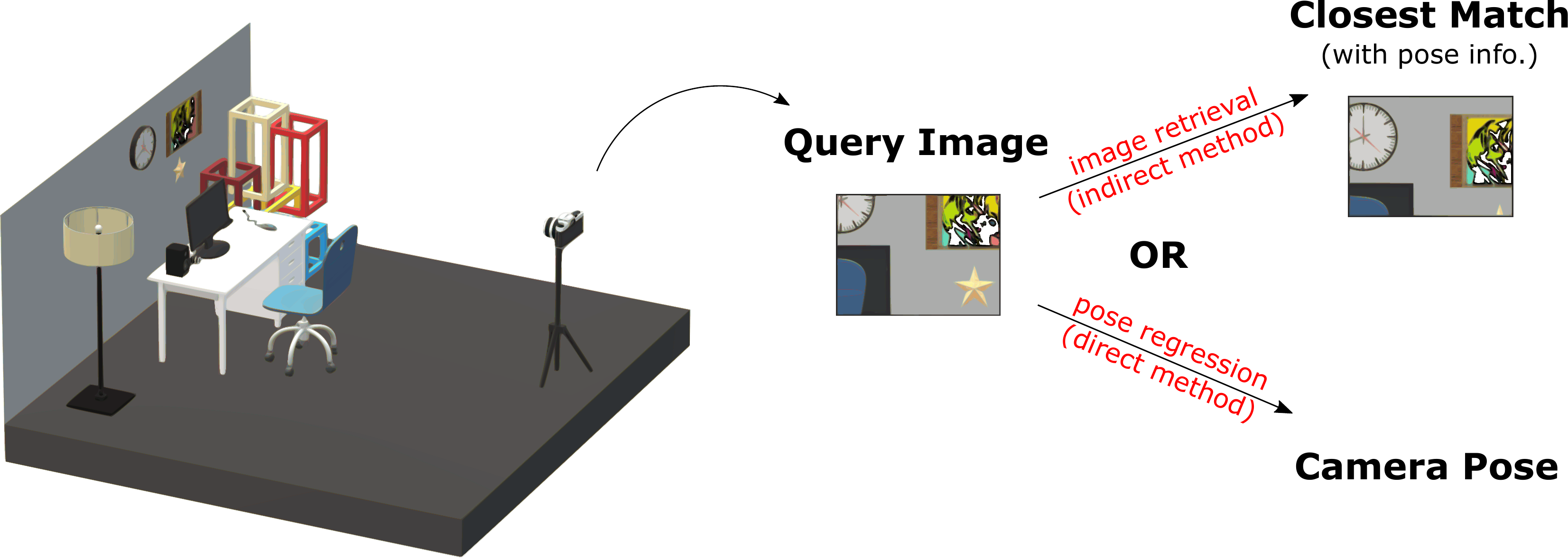}
\caption{The two main approaches to image-based indoor positioning (i.e., indirect and direct).}
\label{direct_vs_indirect}
\end{figure}

As seen in Fig. \ref{direct_vs_indirect}, the methods used for image-based localization can be generally divided into \textit{indirect} and \textit{direct} methods \cite{PIASCO201890}. Indirect methods cast the localization problem as an image retrieval task in which the query image is matched against previously collected images, thus, providing coarse pose information (i.e., position and orientation of the camera). Direct methods, on the other hand, treat the localization problem as a regression task where camera pose is directly estimated from a query image. The main source of positioning error is caused by \textit{perceptual aliasing} \cite{7339473}, in which two images of two different places appear similar due to lighting conditions or repetitive structures and surfaces. To alleviate this issue, many solutions rely on classical feature-detection algorithms such as Scale-Invariant Feature Transform (SIFT), Affine-SIFT, and Speeded Up Robust Features (SURF) to extract robust, invariant features \cite{Kawaj2010, 6071954, LiandJinling, 7577134}. While powerful, such algorithms are computationally expensive and require the additional step of feature-matching, instigating positioning latencies in the order of seconds if not minutes \cite{Kawaj2010, qu2016evaluation, kendall2015posenet}.

One of the earliest attempts of image-based indoor positioning was conducted by Starner \textit{et al}. in \num{1998} \cite{729529}. The images captured by two hat-mounted cameras, one facing forward and the other downward, were used for positioning by employing a Hidden Markov Model (HMM) to model a user transitioning between adjacent rooms. Primitive features were used, composed of the mean value of the red, green, blue, and luminance pixels. A room classification accuracy of \SI{82}{\percent} was achieved inside a \num{14}-room testbed.

\begin{figure}[!b]
\centering
\includegraphics[width=0.65\columnwidth]{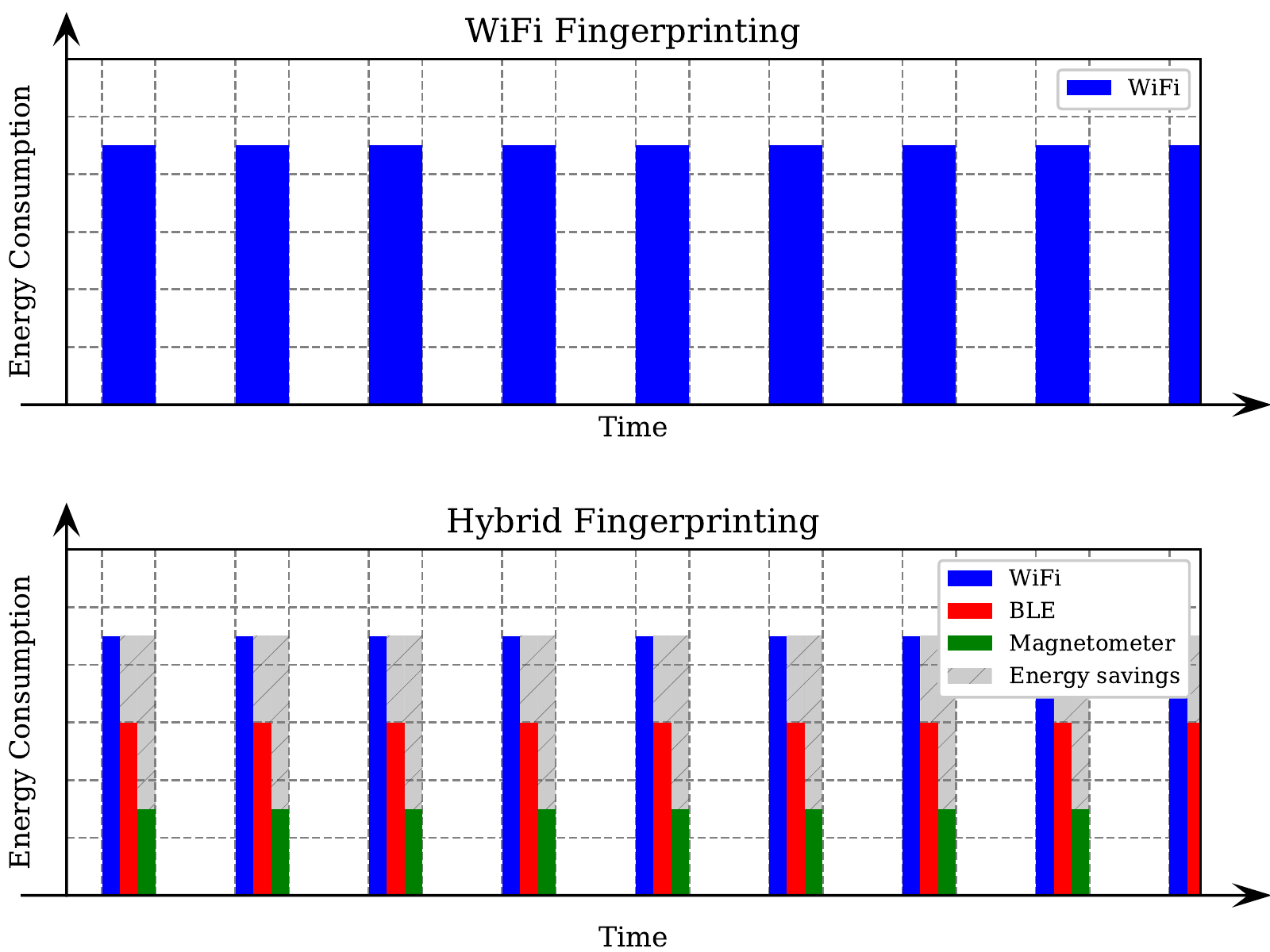}
\caption{An illustration of how hybrid fingerprints can reduce energy consumption. The upper plot represents a system that uses WiFi-only fingerprints, while the lower plot represents a system that uses a combination of WiFi, BLE, and magnetic field fingerprints. The scan rate/period is the same for both systems.}
\label{hybrid}
\end{figure}

\subsection{Hybrid Fingerprints}

A hybrid fingerprinting system is a system that utilizes two or more fingerprint types for positioning. Hybrid fingerprinting systems aim to improve overall performance which can take the form of:
\begin{enumerate}
\item \textit{Improved accuracy}: Combining different fingerprint types provides additional location-specific information. It increases feature dimensionality, resulting in a richer feature set that, in turn, enhances location discrimination. This is often demonstrated in literature by quantifying the gain in positioning accuracy obtained by using multimodal fingerprints instead of unimodal fingerprints \cite{Azizyan_2009}. Nonetheless, cautious handling of sensor synchronization and data fusion is essential to minimize the impact on response time \cite{5651783}.
\item \textit{Improved energy efficiency}: Since different sensors vary in their power requirements, low-power sensors can be exploited to enhance the energy efficiency of an otherwise less-efficient system. This concept is visually illustrated in Fig. \ref{hybrid} however, this requires optimal \textit{sensor scheduling} since degradation in positioning accuracy is expected if the time allocated for WiFi/BLE scanning isn’t enough to detect all APs/beacons necessary for positioning \cite{he2016wi}. Another way of enhancing energy efficiency is to activate sensors only when needed. To help decide when to activate/deactivate sensors, IMU and other sensor measurements can be analyzed to identify a user’s state (stationary vs. walking) \cite{5722356}, as well as a phone’s state (handheld vs. in-pocket) \cite{miluzzo2010pocket}.
\item \textit{Improved availability}: Hybrid fingerprints form the basis for \textit{opportunistic localization} \cite{7874080}. The idea of opportunistic localization is to maximize a system’s availability through the exploitation of all available fingerprint types in a given environment, without relying on specific infrastructure. It can be viewed as a fallback solution in case some fingerprint types cannot be obtained due to infrastructure maintenance/failure. The main drawback of opportunistic localization is its high implementation complexity.
\end{enumerate}

\textit{SurroundSense}, proposed by Azizyan \textit{et al}. in \num{2009} \cite{Azizyan_2009}, is recognized by many as the first hybrid fingerprinting system. The system combines multiple fingerprint types, such as sound, visible light, WiFi, and image fingerprints, to increase location discernibility. Evaluation results across \num{51} stores/shops demonstrated the system’s ability to provide symbolic positioning with \SI{87}{\percent} accuracy. This is an increase of \SI{24}{\percent}, \SI{17}{\percent}, and \SI{13}{\percent} in positioning accuracy over WiFi, sound-and-WiFi, and sound-light-image fingerprints, respectively. However, the system's design is very complicated because it involves several filtering, formatting, matching, clustering, and audio/image processing modules.

\subsection{Miscellaneous Fingerprints}
\subsection*{1) UWB Fingerprints}
Ultra-Wide Band (UWB) is a wireless technology designed for high-bandwidth, short-range (\textless \SI{10}{\meter}) communication. It works by transmitting ultra-short pulses (\textless \SI{1}{\nano\second}) across a wide spectrum of frequency bands (\textgreater \SI{500}{\mega\hertz}). Although the FCC permitted the operation of UWB in \num{2002} \cite{FCC_UWB}, slow progress in standardizing the technology has limited its adoption in consumer devices \cite{1657900}. Concerning indoor positioning, UWB has proved superior to other wireless technologies, specifically for lateration-based approaches, due to its high time delay resolution and, hence, multipath resilience \cite{BonenbergandJames}.

\subsection*{2) Visible Light Fingerprints}
The emergence of Visible Light Communication (VLC) recently enabled Light Emitting Diode (LED)-based indoor positioning \cite{luo2017indoor}. Due to the high directivity of visible light, LED-based positioning systems can provide sub-meter accuracy (based on lateration/angulation) \cite{luo2017indoor}. Moreover, LEDs are low-cost, energy-efficient, provide stable performance, and have a long lifetime ($\sim$\si{50,000} hours). However, one drawback is the degradation of performance in NLoS conditions since VLC is inherently an LoS technology. Also, the coverage of such systems is low because visible light cannot penetrate opaque objects such as walls and panel partitions. Also, in green buildings, where, during the day, lighting is provided by sunlight, an LED-based positioning system may not be a practicable solution.

\subsection*{3) RFID Fingerprints}
Radio Frequency IDentification (RFID) is a wireless technology designed to retrieve data from transponders in proximity. Unlike WiFi or Bluetooth, RFID is not supported on mobile devices. Thus, RFID-based applications assume the deployment of dedicated infrastructure (RFID  readers and tags). This makes RFID an unappealing and costly option for positioning. Nevertheless, due to their energy-efficient and durable operation, RFID has been widely used for asset management and access control \cite{7915680}.

\subsection*{4) Acoustic Fingerprints}
The least popular indoor positioning systems are acoustic-based. This is due to the many challenges that arise when using acoustic signals for indoor positioning such as the strong attenuation of aerial acoustic signals, the limited bandwidth of microphones, the various interferences in the audible band, the short operation distance, and the associated sound pollution \cite{Liu2013Guoguo}. Nevertheless, given how water, as a propagation medium, favors acoustic over radio frequency and light signals, acoustic signals are widely used for underwater positioning \cite{TAN20111663}. 

\section{Indoor Positioning Datasets}
\label{sec3}
This section provides a detailed review of datasets that are used to develop and benchmark fingerprinting systems. The datasets were selected based on various criteria, the most important of which was their suitability for training deep learning models from scratch. Deep learning is inherently a data-intensive endeavor. In other words, one of the major drawbacks of deep learning is its need for large datasets for training. Therefore, a dataset must at least contain thousands of location-tagged instances to qualify for review. Small-scale datasets, such as those described in \cite{toth2016miskolc, popleteev2017ambiloc, cramariuc2016open, walch2017image} were omitted from this review. However, small-scale datasets can be used to fine-tune pre-trained models as demonstrated in \cite{walch2017image}. Other selection criteria included scientific quality, novelty, and potential application domains. Eleven datasets were identified and categorized into four categories according to the data types that they represent: radio frequency, magnetic field, image, and hybrid. 

The first category, radio frequency, comprises four datasets of RSS fingerprints collected from either off-the-shelf smart devices or custom-built devices. The second category, magnetic field, contains two datasets of annotated magnetic field and IMU measurements captured using smartphones. The third category, image, contains two datasets of image fingerprints with accurate and precise position and pose information. The fourth category, hybrid, includes three labeled datasets of heterogeneous data simultaneously recorded using the same smart devices. The datasets within each group are described in ascending order by publication date. Table \ref{dataset_comp_table1} provides a side-by-side comparison of all discussed datasets with respect to the collection environment, while Table \ref{dataset_comp_table2} compares the datasets with respect to the sampling nature and collection platform. Table \ref{dataset_pros_cons_table} highlights some of the datasets' pros and cons and provides the download link for each dataset.

\begin{scriptsize}
\begin{table*}[!t]
\tiny
\renewcommand{\arraystretch}{1.3}
\caption{a side-by-side comparison of the datasets with respect to the collection environment}
\label{dataset_comp_table1}
\centering
\begin{tabular}{|m{2.1cm}|m{3.8cm}|m{1.1cm}|m{1.1cm}|m{1.1cm}|m{1.1cm}|m{1.1cm}|m{1.1cm}|m{1.5cm}|}
\hline
\bfseries Dataset (Year) & \bfseries Type & \bfseries Buildings& \bfseries Floors & \bfseries Rooms & \bfseries Corridors& \bfseries Area (\SI{}{\meter\squared}) & \bfseries RPs & \bfseries Spacing of RPs (\SI{}{\meter})\\
\hline
\multicolumn{9}{c}{Radio  Frequency}\\
\hline
 UJIIndoorLoc (\num{2014}) & University buildings &3&13&254&-&108,703&933&-  \\
\hline

\cite{data3010003} (\num{2018})& University library & 1 & 2& - & - & 432 & 212 & - \\
\hline

\cite{byrne2018residential} (\num{2018})& Residential homes&4&7&34&-&350&194&1  \\
\hline

\cite{s18124462} (\num{2018}) & A research facility &1&1&8&1&237&277&0.6\\
\hline
\multicolumn{9}{c}{Magnetic Field and IMU}\\ 
\hline
UJIIndoorLoc-Mag (\num{2015})& A research lab &1&1&1&8&260&-&- \\
\hline
MagPIE  (\num{2017})&  University buildings &3&3&-&-&960&-&-\\
\hline
\multicolumn{9}{c}{Image}\\ 
\hline
\num{7}-Scenes (\num{2013})& An office space & 1&1&7&-&36.5&-&- \\
\hline
 Warehouse (\num{2018}) &A warehouse&1&1&-&-&875&-&- \\
\hline
\multicolumn{9}{c}{Hybrid}\\ 
\hline
\cite{barsocchi2016multisource} (\num{2016})& A research facility&1&1&3&3&185&325&0.6  \\
\hline
PerfLoc (\num{2016})&  Office; Industrial warehouses; Subterranean structure &4&7&-&-&30,000&900+&-\\
\hline
\cite{8612930} (\num{2019}) &- &1&1&4&2&651&70&- \\
\hline

\end{tabular}
\end{table*}
\end{scriptsize}

\begin{scriptsize}
\begin{table*}[!t]
\tiny
\renewcommand{\arraystretch}{1.3}
\caption{a side-by-side comparison of the datasets with respect to the sampling nature and the collection platform}
\label{dataset_comp_table2}
\centering
\begin{tabular}{|m{2.1cm}|m{1.5cm}|m{1.1cm}|m{0.8cm}|m{0.6cm}|m{0.6cm}|m{1.1cm}|m{0.5cm}|m{1.9cm}|m{0.6cm}|m{2.3cm}|}
\cline{2-11}

\multicolumn{1}{c|}{}&\multicolumn{5}{c|}{\bfseries Samples} &  \multicolumn{5}{c|}{\bfseries Platform}\\
\hline

\bfseries Dataset (Year) & \bfseries Type & \bfseries Rate (\SI{}{\hertz})& \bfseries Training & \bfseries Testing & \bfseries Features & \bfseries Collection side & \bfseries Devices & \bfseries Type & \bfseries OS & \bfseries Orientation\\
\hline
\multicolumn{11}{c}{Radio  Frequency}\\
\hline
UJIIndoorLoc (\num{2014})&Discrete&-&19,938&1,111&520&User&25&Smartphone; Tablet&Android&Not provided  \\
\hline

\cite{data3010003} (\num{2018}) &  Discrete&-&$\sim$15,500&$\sim$88,000&620&User&1&Smartphone&Android& Provided for only two directions \\
\hline

\cite{byrne2018residential} (\num{2018}) & Discrete; Continuous& 5; 25 & $\sim$730,000 &-&varies&Nodes&8 or 11&Raspberry Pi&-&Provided \\
\hline

\cite{s18124462} (\num{2018}) & Discrete; Continuous&10&$\sim$2,820,000&-&varies&User; Nodes&1 to 11&Raspberry Pi; Smartphone&Android&Provided  \\
\hline
\multicolumn{11}{c}{Magnetic Field and IMU}\\ 
\hline
UJIIndoorLoc-Mag (\num{2015})& Continuous&10&270&11&9&User&2&Smartphone&Android&Provided \\
\hline
MagPIE  (\num{2017}) & Continuous&50; 200&591&132&9&User&2&Smartphone&Android&Provided \\
\hline
\multicolumn{11}{c}{Image}\\ 
\hline
\num{7}-Scenes (\num{2013})& Discrete; Continuous&-&26,000&17,000&307,200&User&1&Kinect RGB-D camera&-&Provided \\
\hline
Warehouse (\num{2018}) & Discrete; Continuous&-&202,224&262,570&307,200&User&8&Web camera&-&Provided \\
\hline
\multicolumn{11}{c}{Hybrid}\\ 
\hline
\cite{barsocchi2016multisource} (\num{2016}) & Discrete &10&36,795&-&varies&User&2&Smartphone; Smartwatch&Android&Provided\\
\hline
PerfLoc (\num{2016}) & Discrete; Continuous& from 0.3 to 100&varies&private&varies&User&4&Smartphone&Android&Provided\\
\hline

\cite{8612930} (\num{2019}) & Discrete &-&1,010,640&-&16&Nodes&5&Raspberry Pi&-&Provided for only one angle   \\
\hline

\end{tabular}
\end{table*}
\end{scriptsize}

\begin{scriptsize}
\begin{table*}[!t]
\tiny
\renewcommand{\arraystretch}{1.3}
\caption{the pros, cons, and download link for each dataset}
\label{dataset_pros_cons_table}
\centering
\begin{tabular}{|m{1.5cm}|m{5.5cm}|m{6.4cm}|m{2.9cm}|}
\hline
\bfseries Dataset (Year) & \bfseries Pros & \bfseries Cons& \bfseries Download Link \\
\hline
UJIIndoorLoc (\num{2014}) & Unique in terms of the area covered, the number of RPs surveyed, and the number of devices used in data collection. & No orientation information was provided which may lead to inconsistent measurements \cite{zanella2016best}.  & \url{https://archive.ics.uci.edu/ml/datasets/ujiindoorloc} \\
\hline
\cite{data3010003} (\num{2018})& Samples were collected over \num{25} months which helps study temporal signal variations for the development of systems robust to these variations.& Samples were collected facing only two opposing direction for each RP. Didn't specify whether environment changes have occurred during the collection period. & \url{https://doi.org/10.5281/zenodo.1309317} \\
\hline

\cite{byrne2018residential} (\num{2018})& Since data were collected from private residential homes and from various activity zones, it is appealing for studying indoor tracking in support of AAL. & Not suited for studying smartphone-based indoor positioning. & \url{https://doi.org/10.6084/m9.figshare.6051794.v5} \\
\hline

\cite{s18124462} (\num{2018}) & Data was collected from both user and node sides. Various scenarios and transmission powers were explored. & The samples corresponding to a user/node sending signals to itself were not filtered out. & \url{http://wnlab.isti.cnr.it/localization}\\
\hline

UJIIndoorLoc-Mag (\num{2015})& Data collection was repeated several times over the same path which makes it easier to detect noise and outliers in the measurements. & Provides very few calibration points since ground truth location information was only recorded at the beginning and end of each line segment. & \url{http://archive.ics.uci.edu/ml/datasets/UJIIndoorLoc-mag} \\
\hline
MagPIE  (\num{2017})& Data were collected with and without the placement of live loads. Orientation of the smartphone kept fixed throughout which is key for consistent magnetic field measurements. & Relied on Google Tango for ground truth measurements which has proven to be an unreliable source for accurate measurements \cite{roberto2016evaluation}. & \url{http://bretl.csl.illinois.edu/magpie/} \\
\hline
\num{7}-Scenes (\num{2013})& Includes depth images which is compelling as smartphones equipped with depth cameras have recently started to appear in the market. & Each room has its own coordinate system which is contrary to real life scenarios in which an indoor environment composed of multiple rooms share the same coordinate system. & \url{https://www.microsoft.com/en-us/research/project/rgb-d-dataset-7-scenes/} \\
\hline
Warehouse (\num{2018})& Various testing scenarios and highly accurate and precise ground truth measurements. & Requires more than 30 GB of memory space to store the entire dataset. & \url{https://www.iis.fraunhofer.de/warehouse} \\
\hline

\cite{barsocchi2016multisource} (\num{2016})& Contains samples collected from a smartwatch. Additionally, magnetic field data was collected from rooms rather than corridors only. & The arrival and departure timestamps of some RPs are missing and the WiFi fingerprints were collected from the smartphone only. & \url{http://wnet.isti.cnr.it/software/Ipin2016Dataset.html} \\
\hline
PerfLoc (\num{2016})& Most diversified in terms of the data types collected. Moreover, data were collected to comply with most of the testing and evaluation criteria as specified by the ISO/IEC 18305:2016 standard. & Non-uniform sampling rates across smartphones resulted in asynchronous data samples. Also, data is not directly accessible as there is a steep learning curve to decode the data before start using it \cite{moayeri2018perfloc}. & \url{https://perfloc.nist.gov/} \\
\hline
\cite{8612930} (\num{2019}) & Well-suited for studying indoor tracking using hybrid measurements. Moreover, the dataset contains Xbee measurements and has over \num{1} million samples.  & Orientation is provided around a single axis only (i.e., yaw/heading angle). Not suited for studying smartphone-based indoor positioning.& \url{http://www.gatv.ssr.upm.es/~abh/} \\
\hline

\end{tabular}
\end{table*}
\end{scriptsize}

\subsection{Radio Frequency Datasets}

\subsection*{UJIIndoorLoc:}
The UJIIndoorLoc dataset \cite{torres2014ujiindoorloc}, proposed in \num{2014}, is well known for being the first publicly available RSS dataset. It was created to address the lack of a common dataset for comparing state-of-the-art WiFi fingerprinting systems. The data were collected from three adjacent multi-floor buildings (\num{4}-\num{5} floors) of the Jaume I University campus. A single RP was placed at the center of each room and in front of the door(s) leading to the rooms. \num{25} smart devices carried by \num{20} participants were used to collect over \si{20,000} discrete samples from \num{933} RPs. Each sample is comprised of \num{520} RSS measurements corresponding to the \num{520} APs scattered across the buildings along with ground truth information, such as building and floor numbers, latitude and longitude, a timestamp, and user and device labels. The RSS value of a detected AP ranged from \si{0}{dBm} (very strong signal) to \si{-104}{dBm} (very weak signal). Undetected APs were given an artificial value of \si{+100}{dBm}. On average, \num{27} APs were detected per RP. \SI{5}{\percent} of the collected samples were dedicated as a separate testing set. The authors provided a baseline of an \SI{89.92}{\percent} hit rate and a \SI{7.9}{\meter} mean error using the $k$NN classifier (with $k=1$ and a Euclidean distance metric).

\subsection*{Dataset described in \cite{data3010003}:}
The dataset described in \cite{data3010003} was collected over fifteen months. The primary goal of creating the dataset was to provide researchers with the data needed to study a system's robustness against short/long-term WiFi signal variations. Short-term variations are caused by multipath and shadowing while long-term variations are caused by environment and network changes. Data was collected using a smartphone on two identical floors (\num{3}\textsuperscript{rd}and \num{5}\textsuperscript{th}) of a \num{12}$\times$\SI{18}{\meter\squared} library wing with \num{106} RPs per floor. At each RP, consecutive samples facing the same directions were collected, multiple times a month. During a month,  \SI{15}{\percent} of the samples collected were allocated for training while the remaining \SI{85}{\percent} were allocated for testing, except for the samples collected during the first month (\SI{73}{\percent} training and \SI{27}{\percent} testing). A total of \si{63,504} samples were collected by last month. Each sample consisted of a timestamp, ground truth floor number, RP coordinates, and the RSS values of all detected APs over the entire period (i.e., starting with \num{77} APs at month \num{1} and ending with \num{448} APs at month \num{15}). Recently, the authors updated the dataset to include \si{40,080} new samples corresponding to an additional collection period of ten months with \num{172} newly detected APs. Supporting scripts in MATLAB, that allow for loading a desired set based on filtering criteria, are provided.

\subsection*{Dataset described in \cite{byrne2018residential}:}
The dataset by Byrne \textit{et al}. \cite{byrne2018residential} contains approximately fourteen hours of annotated wearable measurements acquired from four single- and two-floor residential homes with four to eleven rooms. At each residence, a custom-built, wrist-worn transmitter sent accelerometer measurements, via BLE radio (in advertising mode), which were then received by several custom-built anchor nodes deployed throughout the residence. Upon reception, each node records the RSS of the advertised packet and timestamps it. Ground truth location labels were provided through fiducial floor tags that were placed one meter apart throughout the home. A downward-facing camera, strapped to a participant's navel area, automatically captured the floor tags as the participant traversed them. At each floor tag, data were collected facing each of the four cardinal directions to account for the shadowing effect imposed by the participant's body. Additionally, the dataset incorporated samples generated from both scripted and unscripted scenarios. Scripted scenarios represented walking rapidly or slowly throughout the residence while unscripted scenarios represented participants carrying out their normal daily living routine. The dataset also contains annotated data collected from ``activity zones'' (i.e., certain locations coincide with certain activities, such as cooking in the kitchen, eating at the dining table, or relaxing on the sofa). In total, the dataset contains around \si{730,000} samples. Python scripts for loading the dataset form the repository are provided. 

\subsection*{Dataset described in \cite{s18124462}:}
The dataset by Baronti \textit{et al}. \cite{s18124462} was introduced as a general-purpose dataset that can be used for positioning, tracking, proximity/occupancy detection, and social interaction detection. Data collection was performed inside a \num{16.6}$\times$\SI{14.3}{\meter\squared} research facility consisting of eight rooms, a connecting corridor, and \num{277} RPs spaced \SI{0.6}{\meter} apart. Each room contained a Raspberry Pi equipped with two BLE modules. One module continuously listened for signals while the other transmitted advertisements at \SI{10}{\hertz}. Similarly, mobile users carrying a smartphone (as a receiver) and a BLE tag (as a transmitter) were employed to enable data collection both ways (i.e., from user to anchor nodes and vice versa). Six scenarios were used for data collection: ``survey’’, ``localization’’, and four ``social’’. In the survey scenario, the user stood over each RP and collected data along the $+x$, $+y$, $-x$, and $-y$ directions. The localization scenario represented a user walking a predefined path (i.e., continuous sampling). The social scenarios represented two/three users walking from their offices, attending meetings, and returning to their offices. For each scenario, three runs of data collection were performed, corresponding to three transmission powers (i.e., \si{3}{dBm}, \si{-6}{dBm}, and \si{-18}{dBm}). Each sample consists of a timestamp, transmitter ID, receiver ID, and RSS value. Ground truth location information is provided through a separate file that maps timestamps to the coordinates of the RPs. Overall, the dataset has around \si{2,820,000} samples.

\subsection{Magnetic Field Datasets}
\subsection*{UJIIndoorLoc-Mag:}
The creators of the UJIIndoorLoc dataset introduced the UJIIndoorLoc-Mag dataset in \num{2015} \cite{torres2015ujiindoorloc}. The aim was to provide a common dataset for the evaluation of magnetic field fingerprinting systems as they became increasingly popular. Unlike UJIIndoorLoc, the data contained in UJIIndoorLoc-Mag was collected in a much smaller area (a single \num{15}$\times$\SI{20}{\meter\squared} office space). A smartphone was used to collect continuous samples along the office's eight corridors at a sampling rate of \SI{10}{\hertz}. Each continuous sample represents walking along a predefined path composed of multiple straight-line segments. The data collection process involved several predefined paths where sampling over each path was repeated multiple times yielding a total number of \num{281} continuous samples (or \si{40,159} discrete captures). Each discrete capture incorporated timestamped, raw measurements from the phone's magnetometer, accelerometer, and orientation sensor along its three axes (Fig. \ref{phone}). Ground truth location information was recorded at the beginning and end of each continuous sample and turning points (i.e., the end of a segment and the beginning of another). The authors used a subset of the dataset to provide a baseline of a \SI{7.23}{\meter} mean error using the $k$NN classifier (with $k=1$ and a Euclidean distance metric).

\subsection*{MagPIE:}
The Magnetic Positioning Indoor Estimation (MagPIE) dataset \cite{hanley2017magpie} is, by far, the largest dataset for studying and comparing approaches to magnetic and inertial indoor positioning. The data were collected from three different university buildings. A smartphone, either handheld or mounted on a wheeled robot, was used to collect \num{723} continuous samples equaling \SI{51}{\kilo\metre} of total distance traveled. The sampling rate was \SI{50}{\hertz} for magnetometer data and \SI{200}{\hertz} for accelerometer and gyroscope data. To account for soft/hard iron biases, the dataset provides calibrated measurements as opposed to raw magnetic field measurements. A separate smartphone was used to provide ground truth location information by running Google Tango, an augmented reality platform for mobile devices (discontinued March \num{2018}). Data were collected under two scenarios (i.e., with and without the placement of ``live loads''). Live loads are certain objects, commonly found inside buildings, that may affect the magnetometer's measurements. However, the number of live loads placed, their description, and their ground truth location information were not provided. 

\subsection{Image Datasets}
\subsection*{\num{7}-Scenes:}
The \num{7}-Scenes dataset, introduced by Microsoft Research in \num{2013} \cite{Shotton_2013_CVPR}, has been widely used for image-based localization. It is composed of Red-Green-Blue images and their corresponding depth images (collectively called RGB-D images) of seven small-scale indoor scenes. Each scene typically consists of a single room (e.g., office, kitchen). The spatial volume of these scenes ranges from \SI{2}{\meter}$\times$\SI{0.5}{\meter}$\times$\SI{1}{\meter}
to \SI{4}{\meter}$\times$\SI{3}{\meter}$\times$\SI{1.5}{\meter}. All images were captured using a handheld \textit{Kinect} RGB-D camera at \num{640}$\times$\num{480} resolution. Ground truth position and orientation information was provided by the SLAM-based \textit{KinectFusion} system. The number of training images for each scene ranges from \si{1,000} to \si{7,000} while the number of testing images ranges from \si{1,000} to \si{5,000}. Overall, the dataset contains \si{26,000} training images and \si{17,000} testing images. The dataset is considered challenging for positioning algorithms due to notable motion blur, variations in camera pose, and because scenes contain many ambiguous texture-less features.

\subsection*{Warehouse:}
Warehouse \cite{loffler2018evaluation} is a dataset created for the development and benchmarking of image-based localization systems in industrial settings. For data collection, the authors utilized eight web cameras mounted on special platforms that placed them at \num{45}$^\circ$ increments. Each camera captured RGB images at \num{640}$\times$\num{480} resolution inside a \num{25}$\times$\SI{35}{\meter\squared} industrial warehouse. Each image is labeled with a sub-millimeter position and sub-degree orientation information using a laser-based reference system. Two trajectories, intended to uniformly cover the area, were followed to obtain over \si{200,000} training images. The testing images were collected over carefully designed trajectories aimed at evaluating different aspects of the positioning system such as its ability to generalize and respond to environmental changes and scaling and its robustness to local and global ambiguity. The authors provided baselines of \SI{1.08}{\meter} to \SI{6.76}{\meter} mean errors (depending on the testing trajectory) using the CNN-based, pre-trained \textit{PoseNet} \cite{kendall2015posenet}.

\subsection{Hybrid Datasets}
\subsection*{Dataset described in \cite{barsocchi2016multisource}:}
Barsocchi \textit{et al}. \cite{barsocchi2016multisource} collected WiFi, magnetometer, and IMU data from an indoor environment composed of three rooms of different sizes and three corridors of different lengths. Data collection was performed by concurrently wearing two synchronized smart devices: a smartphone and a smartwatch. A fixed sampling rate of \SI{10}{\hertz} was used for both devices. The smartphone was held at chest-level of the person collecting the data, with the screen facing up, while the smartwatch was wrist-worn. Data were collected over two campaigns from \num{325} uniformly distributed and regularly spaced RPs covering a surface area of \SI{185}{\meter\squared}. The ground truth coordinates of these points, along with arrival and departure timestamps at each point, are included in the dataset. In total, the dataset contains over \si{36,000} discrete instances.

\subsection*{PerfLoc:}
For PerfLoc \cite{moayeri2016perfloc}, data were collected based on guidance from the \textit{ISO/IEC 18305:2016} international standard for testing and evaluating Localization and Tracking Systems (LTSs) \cite{ISOIEC201618305}. The standard specifies that localization systems should be evaluated under different environmental and mobility settings. Hence, the data includes timestamped samples collected from four different buildings (including a subterranean structure) using different mobility modes such as walking, running, walking backward, crawling, and sidestepping. Four Android-based smartphones, strapped to the upper arms of the person collecting the data, were employed to collect data from the \num{900}+ RPs placed throughout the buildings. Diverse data were collected including:  WiFi, cellular, GPS, and all other available sensor data for a given smartphone (e.g., magnetic field, acceleration, temperature, pressure, humidity, light intensity, etc.). The sampling rate ranged from \SI{0.3}{\hertz} to \SI{100}{\hertz}, depending on the data type sampled and the smartphone's brand and model. The authors provide a private testing set through an online web portal where developers can upload their location estimates and get real-time feedback on their system's performance.

\subsection*{Dataset described in \cite{8612930}:}
The dataset by Belmonte-Hernández \textit{et al}. \cite{8612930} contains Xbee, BLE, WiFi, and orientation measurements collected in a \si{31}$\times$\SI{21}{\meter\squared} area comprised of four rooms and two corridors. The data were collected using five Raspberry Pi receivers that were strategically placed in the environment. The entire environment was divided into seventy rectangular cells of different sizes, ranging from \num{1.5}$\times$\SI{1.42}{\meter\squared} to \num{2.56}$\times$\SI{1.9}{\meter\squared}. At least five minutes of measurement was recorded for each cell in all \num{360} degrees. A person wearing a Raspberry Pi transmitter attached to their hip would stand at the center of cells to complete data collection. These received measurements were then synchronized and labeled with the coordinates of the cells' centers. Overall, the dataset has about one million samples.

\section{Fingerprinting Evaluation Metrics}
\label{sec4}
A major challenge in indoor positioning is the lack of a universal evaluation framework to fairly compare the performance of different indoor positioning systems \cite{harle2013survey, subbu2014analysis, davidson2017survey}. This is primarily owed to the subtle nature of indoor positioning in general and fingerprinting in particular. For example, the performance of WiFi fingerprinting systems is affected by a plethora of variables \cite{KAEMARUNGSI2012292,6761255}, including:
\begin{itemize}
\item Hardware: orientation, directionality, and type of wireless network interface card (WNIC),
\item Spatial: the distance between receiver and AP,
\item Temporal: time and period of measurement,
\item Interference: radio frequency interference caused by other devices,
\item Human: user's presence, orientation, mobility,
\item Environment: building types and construction materials.
\end{itemize}
Researchers often end up comparing their results to other self-reported results because it is too much of an effort to reproduce implementations for the sake of comparison. This is especially true if the implementation compared against is ill-defined due to a lack of complete disclosure of materials and methods, both of which are essential for reproducibility.

Nonetheless, the proposal of public datasets and the organization of indoor positioning competitions, such as the IPIN \cite{IPIN} and Microsoft \cite{MicrosoftCompetition} competitions, are significant steps towards overcoming these limitations. In addition to acting as neutral grounds for comparison, such initiatives help researchers avoid the costs required for the setup and maintenance of experimental testbeds. A recent attempt to define standards for localization is the proposal of the ISO/IEC 18305:2016 standard \cite{ISOIEC201618305}. The standard defines various Test and Evaluation (T\&E) procedures for LTSs with an emphasis on fire-fighter scenarios. While valuable, the standard has several shortcomings as pointed out by the International Standards Committee of IPIN \cite{potorti2018evaluation}. Also, the standard is useful for end users, but not fully useful for either developers or researchers \cite{8825820}. Since there seems to be no consensus among the research community, how to define benchmarks, evaluation metrics, and standards is still open to interpretation.

This section sheds light on five evaluation metrics that are necessary for characterizing deep learning-based fingerprinting systems. These are accuracy, precision, complexity, cost, and scalability. These metrics constitute an evaluation framework that will be used in the next section to qualitatively compare systems. Note, however, that the metrics do not measure mutually exclusive properties since explicit/implicit correlations between them generally exist. For example, the cost of site surveying and the accuracy of a system is controlled by the density of RPs and the density of collected fingerprints per RP; the denser the RPs and fingerprints, the more time and money is spent on site surveying, and the more accurate the system is expected to perform.

\subsection{Accuracy} 
Accuracy is the most reported metric in indoor positioning. It is a quantity that reflects how close a system's measurements are to ground truth measurements. The mean absolute error (MAE), which is the average \num{2}D/\num{3}D Euclidean distance between ground truth and predicted locations, has been widely adopted as an accuracy metric as well as the mean squared error (MSE) and the root mean squared error (RMSE). The classification accuracy, or hit rate, of floors, rooms, and/or RPs  has also been used to reflect the accuracy of a system. Last, in deep learning-based fingerprinting, top-$N$ accuracy has been used as an accuracy metric. In top-$N$ accuracy, a prediction is considered correct if the correct class is among the $N$ highest probable classes predicted by the system. Note that top-\num{1} accuracy is the same as conventional classification accuracy.

\subsection{Precision}
An important metric that relates to accuracy is precision. Precision captures the agreement between a system's estimates over several independent trials. Quantiles of error, which measure the fraction of trials in which a system produced a certain error (e.g., an error of $\leq$ \SI{2.3}{\meter} \SI{95}{\percent} of the time), are often used as a precision metric. Additionally, the minimum and maximum positioning errors produced by a system represent its error bounds. However, a Cumulative Distribution Function (CDF), or a histogram of the error, is more informative since they represent the distribution of error over all trials. The standard deviation ($\sigma$), as well as the variance ($\sigma^2$) of the error, have also been utilized as precision metrics. In deep learning-based fingerprinting, the confusion matrix serves as an error distribution from which precision can be easily extracted. 

\subsection{Complexity}
In indoor positioning, complexity often refers to a system's computational complexity. Since deriving the analytic complexity formula of different indoor positioning systems is usually an arduous task \cite{liu2007survey, yassin2016recent}, other complexity measures are considered instead. For example, in deep learning-based fingerprinting, training and response times, as well as characteristics about the network's parameters (e.g., their number or the memory space they occupy), have all been utilized as computational complexity indicators. Training time must be considered because updating a fingerprint database necessitates re-training the system. Response time is the elapsed time between requesting and receiving a location estimate after training is complete. A system that delegates positioning inference to the server-side must account for the round-trip latency in the response time. Training and response times are hardware-specific assessments. A more independent measure of computational complexity is the number of floating-point operations (FLOPs) that are required to produce a location estimate. Additionally, an important factor to consider is the sample preparation time. Sample preparation time refers to the time it takes to acquire and process a sample for positioning.

\subsection{Cost}
The total cost of a fingerprinting system includes both initial and subsequent costs. Initial costs are one-time expenditures related to the initial establishment of the system. These include infrastructure acquisition/installment and site surveying costs. Systems that exploit existing infrastructure, such as the WiFi APs used for communication, are thus considered cost-effective. Subsequent costs are recurring expenditures that are required to keep the system up and running. Examples of such expenditures include cloud server rental, energy consumption, fingerprint database updates, and infrastructure repair and maintenance.

\subsection{Scalability} 
The scalability of a fingerprinting system describes its ability to grow and manage demand increase. The demand can take the form of an increase in the area to be covered by the system (e.g., from a single floor to multi-floor to multi-building). The coverage area can be expanded by surveying the new areas and deploying additional hardware (e.g., WiFi APs or BLE beacons) as needed. Sometimes the accuracy of a system needs to be improved. Deploying additional APs or beacons will generally improve accuracy because location discernibility will increase with increased features. However, both coverage area expansion and improved accuracy lead to increased installation and maintenance costs. In this paper, the testbed size of a system is used to show the scale in which it can operate. Another form of demand increase that is more challenging to manage is an increase in the entities to be localized such as hundreds or thousands of localization requests in a busy airport or shopping mall. Given the dynamic nature of this increase (i.e., temporary spikes in requests), mechanisms for managing workloads, where resources are expanded on-demand, need to be implemented. A scalable system must also account for device and user heterogeneity since different users use different smartphones and have different heights, walking patterns, and holding preferences.

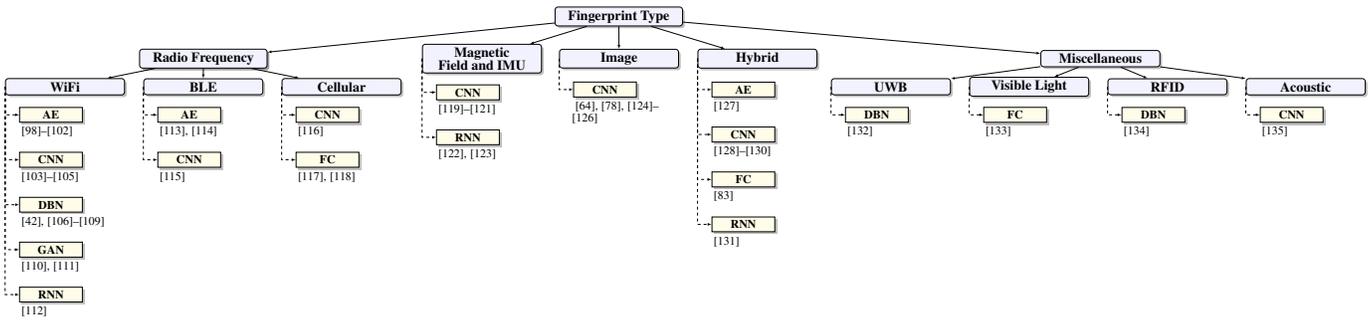
\begin{figure*}[!h]
\centering

\usetikzlibrary{arrows,shapes,positioning,shadows,trees}

\tikzset{
  basic/.style  =  {draw, text width=4cm, drop shadow, rectangle},
  root/.style   =  {basic, thick, text width=4.3cm, fill=blue!5, rounded corners=3pt, align=center},
  level 2/.style = {basic, thick, fill=blue!5, rounded corners=3pt, align=center},
  level 3/.style = {basic, thick, fill=yellow!10, align=center, text width=2cm},
  level 4/.style = {align=left, text width=9.5em}
}

\resizebox{1\textwidth}{!}{%
\begin{tikzpicture}[
  level 1/.style={sibling distance=50mm},
  edge from parent/.style={->,draw},
  >=latex]

% root of the the initial tree, level 1
\node[root] {\Large \bfseries Fingerprint Type}
% The first level, as children of the initial tree
  child {node[level 2, left=2.5cm, style={root}] (c1) {\Large \bfseries Radio Frequency}}
  child {node[level 2, right=-2cm] (c2) {\Large \bfseries Magnetic Field and IMU}}
  child {node[level 2, right=-2.124cm] (c3) {\Large \bfseries Image}}
  child {node[level 2, right=-2.2cm] (c4) {\Large \bfseries Hybrid}}
  child {node[level 2, right=5cm] (c5) {\Large \bfseries Miscellaneous}};

% The second level, relatively positioned nodes
\begin{scope}[every node/.style={level 2}]
\node [below of = c1, xshift=-140pt] (c11) {\Large \bfseries WiFi} ;
\node [right of = c11, xshift=111.5pt] (c12) {\Large \bfseries BLE};
\node [right of = c12, xshift=111.5pt] (c13) {\Large \bfseries Cellular};

\node [below of = c5, xshift=-212pt] (c51) {\Large \bfseries UWB} ;
\node [right of = c51, xshift=111.5pt] (c52) {\Large \bfseries Visible Light};
\node [right of = c52, xshift=111.5pt] (c53) {\Large \bfseries RFID};
\node [right of = c53, xshift=111.5pt] (c54) {\Large \bfseries Acoustic};
\end{scope}

% WiFi --------------
\node [below of = c11, xshift=-0.5cm, style={level 3}] (c111) {\large \bfseries AE};
\node [below of = c111, xshift=0.5cm, yshift=0.4cm, style={level 4}] (c112) {\large \cite{nowicki2017low,8215259,Kim2018,ZHANG2016279,wang2019robust}};

\node [below of = c112, xshift=-0.5cm, style={level 3}] (c113) {\large \bfseries CNN};
\node [below of = c113, xshift=0.5cm, yshift=0.4cm, style={level 4}] (c114) {\large \cite{8027020,CiFi1,CiFi2}};

\node [below of = c114, xshift=-0.5cm, style={level 3}] (c115) {\large \bfseries DBN};
\node [below of = c115, xshift=0.5cm, yshift=0.4cm, style={level 4}] (c116) {\large \cite{DeepFi1,DeepFi2,PhaseFi1,PhaseFi2,BiLoc}};

\node [below of = c116, xshift=-0.5cm, style={level 3}] (c117) {\large \bfseries GAN};
\node [below of = c117, xshift=0.5cm, yshift=0.4cm, style={level 4}] (c118) {\large \cite{8891678,8644149}};

\node [below of = c118, xshift=-0.5cm, style={level 3}] (c119) {\large \bfseries RNN};
\node [below of = c119, xshift=0.5cm, yshift=0.4cm, style={level 4}] (c1110) {\large \cite{8830368}};

% BLE ---------------
\node [below of = c12, xshift=-0.5cm, style={level 3}] (c121) {\large \bfseries AE};
\node [below of = c121, xshift=0.5cm, yshift=0.4cm, style={level 4}] (c122) {\large \cite{7959171,mohammadi2018semisupervised}};

\node [below of = c122, xshift=-0.5cm, style={level 3}] (c123) {\large \bfseries CNN};
\node [below of = c123, xshift=0.5cm, yshift=0.4cm, style={level 4}] (c124) {\large \cite{iqbal2018accurate}};

% Cellular ---------------
\node [below of = c13, xshift=-0.5cm, style={level 3}] (c131) {\large \bfseries CNN};
\node [below of = c131, xshift=0.5cm, yshift=0.4cm, style={level 4}] (c132) {\large \cite{8292280}};

\node [below of = c132, xshift=-0.5cm, style={level 3}] (c133) {\large \bfseries FC};
\node [below of = c133, xshift=0.5cm, yshift=0.4cm, style={level 4}] (c134) {\large \cite{8570849,8446013}};

% Magnetic ---------------
\node [below of = c2, xshift=-0.5cm, yshift=-0.2cm, style={level 3}] (c21) {\large \bfseries CNN};
\node [below of = c21, xshift=0.5cm, yshift=0.4cm, style={level 4}] (c22) {\large \cite{8115887,8626558,8560119}};

\node [below of = c22, xshift=-0.5cm, style={level 3}] (c23) {\large \bfseries RNN};
\node [below of = c23, xshift=0.5cm, yshift=0.4cm, style={level 4}] (c24) {\large \cite{8254556,8660396}};

% Image ---------------
\node [below of = c3, xshift=-0.5cm, yshift=-0.1cm, style={level 3}] (c31) {\large \bfseries CNN};
\node [below of = c31, xshift=0.5cm, yshift=0.2cm, style={level 4}] (c32) {\large \cite{kendall2015posenet,walch2017image,7743683,ha2018image,ACHARYA2019245}};

% Hybrid ---------------
\node [below of = c4, xshift=-0.5cm, yshift=-0.1cm, style={level 3}] (c41) {\large \bfseries AE};
\node [below of = c41, xshift=0.5cm, yshift=0.4cm, style={level 4}] (c42) {\large \cite{8331081}};

\node [below of = c42, xshift=-0.5cm, style={level 3}] (c43) {\large \bfseries CNN};
\node [below of = c43, xshift=0.5cm, yshift=0.4cm, style={level 4}] (c44) {\large \cite{7776928,8354175,8554268}};

\node [below of = c44, xshift=-0.5cm, style={level 3}] (c45) {\large \bfseries FC};
\node [below of = c45, xshift=0.5cm, yshift=0.4cm, style={level 4}] (c46) {\large \cite{8612930}};

\node [below of = c46, xshift=-0.5cm, style={level 3}] (c47) {\large \bfseries RNN};
\node [below of = c47, xshift=0.5cm, yshift=0.4cm, style={level 4}] (c48) {\large \cite{8422562}};

% UWB ---------------
\node [below of = c51, xshift=-0.5cm, style={level 3}] (c511) {\large \bfseries DBN};
\node [below of = c511, xshift=0.5cm, yshift=0.4cm, style={level 4}] (C512) {\large \cite{luo2016deep}};

% Visisble light ---------------
\node [below of = c52, xshift=-0.5cm, style={level 3}] (c521) {\large \bfseries FC};
\node [below of = c521, xshift=0.5cm, yshift=0.4cm, style={level 4}] (c522) {\large \cite{8693950}};

% RFID ---------------
\node [below of = c53, xshift=-0.5cm, style={level 3}] (c531) {\large \bfseries DBN};
\node [below of = c531, xshift=0.5cm, yshift=0.4cm, style={level 4}] (c532) {\large \cite{8761800}};

% Acoustic ---------------
\node [below of = c54, xshift=-0.5cm, style={level 3}] (c541) {\large \bfseries CNN};
\node [below of = c541, xshift=0.5cm, yshift=0.4cm, style={level 4}] (c542) {\large \cite{8543833}};

% lines from each level 1 node to every one of its "children"
\foreach \value in {1,2,3}
  \draw[->] (c1) -- (c1\value);
  
\foreach \value in {1,2,3,4}
  \draw[->] (c5) -- (c5\value);

% WiFi  
\foreach \value in {1,3,5,7,9}
  \draw[dashed,->] (c11.west) |- (c11\value.west);
  
% BLE  
\foreach \value in {1,3}
  \draw[dashed,->] (c12.west) |- (c12\value.west);
  
% Cellular  
\foreach \value in {1,3}
  \draw[dashed,->] (c13.west) |- (c13\value.west);

% Magnetic
\foreach \value in {1,3}
  \draw[dashed,->] (c2.west) |- (c2\value.west);  
  
% Image
\foreach \value in {1}
  \draw[dashed,->] (c3.west) |- (c3\value.west);

% Hybrid
\foreach \value in {1,3,5,7}
  \draw[dashed,->] (c4.west) |- (c4\value.west);
  
% UWB  
\foreach \value in {1}
  \draw[dashed,->] (c51.west) |- (c51\value.west);
  
% Visisble light  
\foreach \value in {1}
  \draw[dashed,->] (c52.west) |- (c52\value.west);

% Acooustic  
\foreach \value in {1}
  \draw[dashed,->] (c53.west) |- (c53\value.west);
  
% RFID  
\foreach \value in {1}
  \draw[dashed,->] (c54.west) |- (c54\value.west);

\end{tikzpicture}
}%
\caption{A two-level taxonomy is followed where solutions are classified based on the fingerprint type and then sub-classified based on the deep learning model.}
\label{taxonomy}
\end{figure*} 

\section{Deep Learning-Based Indoor Positioning Solutions}
\label{sec5}
This section provides an in-depth review of existing fingerprinting solutions based on deep learning methods. As illustrated in Fig. \ref{taxonomy}, these solutions are classified using a two-level taxonomy (i.e., based on the fingerprint type employed and further subdivided based on the deep learning model used). The fingerprint types include Radio Frequency (WiFi, BLE, and Cellular), Magnetic Field and IMU, Image, Hybrid, and Miscellaneous (UWB, Visible Light, RFID, and Acoustic). The deep learning methods include AE, CNN, DBN, FC, GAN, and RNN.

 Table \ref{fingerprints_table_1} summarizes and compares the reviewed solutions based on the performance evaluation metrics discussed in the previous section. The table entries are populated based on information extracted directly from corresponding papers.

\textit{It should be noted that the results presented in the table are not good indicators for comprehending which solution is better than another. These solutions have been implemented in different testbeds that differ in size, number of RPs, the granularity of RPs, number of training/testing samples, etc. To draw safe conclusions, all solutions should be implemented in the same testbed, which is not feasible. Nevertheless, these results still provide valuable insight into how a solution could perform in practical applications or settings}.

\subsection{Radio Frequency Fingerprints}

\subsection*{1) WiFi Fingerprints}

\begin{table*}[!t]
\tiny
\renewcommand{\arraystretch}{1.3}
\caption{a summary and comparison of the reviewed fingerprinting solutions}
\label{fingerprints_table_1}
\centering
\begin{tabular}{|m{0.4cm}|m{1.6cm}|m{1cm}|m{2.6cm}|m{2.7cm}|m{2.1cm}|m{1.6cm}|m{2.6cm}|}
\hline
\bfseries Work and year & \bfseries Purpose & \bfseries Model and framework & \bfseries Accuracy & \bfseries Precision & \bfseries Complexity & \bfseries Cost & \bfseries Scalability\\
\hline
\multicolumn{8}{c}{WiFi Fingerprinting Solutions}\\
\hline
\cite{DeepFi1,DeepFi2} \num{2015} & CSI-amplitude fingerprinting using a single AP& DBN + probabilistic refining  & positioning errors of \SI{0.94}{\meter} and \SI{1.80}{\meter} for LoS and NLoS environments, respectively &  CDF is provided; positioning errors of $\leq$ \SI{1.6}{\meter} and \SI{2.1}{\meter} \SI{80}{\percent} of the time for LoS and NLoS environments, respectively; $\sigma$ of \SI{0.56}{\meter} and \SI{1.34}{\meter} for LoS and NLoS environments, respectively & requires a measurement window of \SI{5}{\second} before computing an output + \SI{2.56}{\second} response time & used \num{1} AP and \num{1} modified WNIC & LoS: a \num{7}$\times$\SI{4}{\meter\squared} empty living room with \num{38} training RPs and \num{12} testing RPs; NLoS: a \num{9}$\times$\SI{6}{\meter\squared} cluttered computer lab with \num{50} training RPs and \num{30} testing RPs; RP spacing in both testbeds is \SI{0.5}{\meter}\\
\hline

\cite{PhaseFi1,PhaseFi2} \num{2015} & CSI-phase fingerprinting using a single AP& DBN + probabilistic refining & positioning errors of \SI{1.08}{\meter} and \SI{2.01}{\meter} for LoS and NLoS environments, respectively & CDF is provided; positioning errors of $\leq$ \SI{1.4}{\meter} and \SI{2.8}{\meter} \SI{80}{\percent} of the time for LoS and NLoS environments, respectively; $\sigma$ of \SI{0.40}{\meter} and \SI{1.01}{\meter} for LoS and NLoS environments, respectively & requires a measurement window of \SI{1}{\second} before computing an output + \SI{0.37}{\second} response time & used \num{1} AP and \num{1} modified WNIC &  LoS: a \num{7}$\times$\SI{4}{\meter\squared} empty living room with \num{38} training RPs and \num{12} testing RPs; NLoS: a \num{9}$\times$\SI{6}{\meter\squared} cluttered computer lab with \num{50} training RPs and \num{30} testing RPs; RP spacing in both testbeds is \SI{0.5}{\meter}\\
\hline

\cite{BiLoc} \num{2017} & CSI-amplitude and AoA fingerprinting in the \SI{5}{\giga\hertz} band & DBN + probabilistic refining & positioning errors of \SI{2.15}{\meter} and \SI{1.57}{\meter} for LoS and NLoS environments, respectively & CDF is provided; positioning errors of $\leq$ \SI{2.8}{\meter} and \SI{2.4}{\meter} \SI{80}{\percent} of the time for LoS and NLoS environments, respectively; $\sigma$ of \SI{1.54}{\meter} and \SI{0.83}{\meter} for LoS and NLoS environments, respectively & requires a measurement window of \SI{0.25}{\second} before computing an output + \SI{0.6}{\second} response time & used \num{2} modified WNICs (one acting as the AP and the other as the mobile node) &  LoS: a \num{2.4}$\times$\SI{24}{\meter\squared} corridor with \num{10} training RPs and \num{10} testing RPs; NLoS: a \num{9}$\times$\SI{6}{\meter\squared} cluttered computer lab with \num{15} training RPs and \num{15} testing RPs; RP spacing in both testbeds is \SI{1.8}{\meter} \\
\hline

\cite{CiFi1,CiFi2} \num{2017} & AoA fingerprinting in the \SI{5}{\giga\hertz} band & CNN + weighted averaging & positioning errors of \SI{2.38}{\meter} and \SI{1.78}{\meter} for LoS and NLoS environments, respectively & CDF is provided; positioning errors of $\leq$ \SI{3.4}{\meter} and \SI{2.4}{\meter} \SI{80}{\percent} of the time for LoS and NLoS environments, respectively; $\sigma$ of \SI{1.45}{\meter} and \SI{1.24}{\meter} for LoS and NLoS environments, respectively & requires a measurement window of \SI{1}{\second} before computing an output + \SI{0.59}{\second} response time & used \num{2} modified WNICs (one acting as the AP and the other as the mobile node) & LoS: a \num{2.4}$\times$\SI{24}{\meter\squared} corridor with \num{10} training RPs and \num{10} testing RPs; NLoS: a \num{9}$\times$\SI{6}{\meter\squared} cluttered computer lab with \num{15} training RPs and \num{15} testing RPs; RP spacing in both testbeds is \SI{1.8}{\meter}; RP spacing in both testbeds is \SI{1.8}{\meter} \\
\hline

\cite{ZHANG2016279} \num{2016} & indoor/outdoor RSS fingerprinting & SDAE + HMM & RMSE of \SI{0.39}{\meter} and \SI{0.36}{\meter} for indoor and outdoor testbeds, respectively & - &  \SI{0.25}{\second} response time for SDAE & used \num{163} and \num{359} already deployed APs for the indoor and outdoor testbeds, respectively &  indoor testbed: a building's floor with \num{91} RPs of \SI{1.8}{\meter} spacing; outdoor testbed: a campus lawn with \num{105} RPs of \SI{1.8}{\meter} spacing \\
\hline

\cite{nowicki2017low} \num{2017} & multi-building and multi-floor classification & SAE + FC; Keras for TensorFlow &  \SI{92}{\percent} classification accuracy & - & - & - & multi-building and multi-floor (UJIIndoorLoc dataset) \\
\hline

\cite{8215259,Kim2018} \num{2017} & scalable fingerprinting for large-scale environments& SAE + FC + weighted centroid; Keras for TensorFlow &  \SI{99.82}{\percent} building hit rate, \SI{91.27}{\percent} floor hit rate, and \SI{9.29}{\meter} positioning error & - & - & - & multi-building, multi-floor, and multi-location (UJIIndoorLoc dataset) \\
\hline

\cite{8027020} \num{2017} &  fingerprinting using CSI images and a single AP& CNN + weighted centroid; Caffe & \SI{1.36}{\meter} positioning error & CDF is provided; positioning errors of $\leq$ \SI{2.5}{\meter} \SI{80}{\percent} of the time; $\sigma$ of \SI{0.90}{\meter} & - & used \num{1} AP and \num{1} modified WNIC & a \num{16.3}$\times$\SI{17.3}{\meter\squared} office space with \num{5} rooms \\
\hline

\cite{8830368} \num{2019} & alleviating location ambiguity by using consecutive RSS fingerprints & LSTM + sliding window averaging  & positioning errors of \SI{0.75}{\meter} and \SI{4.2}{\meter} using the authors' and the UJIIndoorLoc datasets, respectively & CDF is provided; positioning errors of $\leq$ \SI{1.25}{\meter} and \SI{7}{\meter} \SI{80}{\percent} of the time using the authors' and the UJIIndoorLoc datasets, respectively; $\sigma$ of \SI{0.64}{\meter} and \SI{3.2}{\meter} using the authors' and the UJIIndoorLoc datasets, respectively & - & used the existing infrastructure of \num{6} APs; used a wheeled robot equipped with LiDAR for data collection and ground truth measurements & a \si{16}$\times$\SI{21}{\meter\squared} university floor + \num{2} buildings of the UJIIndoorLoc dataset\\
\hline

\cite{8891678,8644149} \num{2019} &  reducing the cost of site surveying by expanding the training set with artificial CSI-amplitude fingerprints & GAN (both generator and discriminator are CNNs) + SVM; TensorFlow & improved positioning accuracy by \SI{11.94}{\percent}, i.e., from a positioning error of \SI{1.34}{\meter} (using real fingerprints only), to a positioning error of \SI{1.18}{\meter} (using real and artificial fingerprints) & CDF is provided; improved precision, i.e., from min./max. positioning errors of \SI{0.12}{\meter}/\SI{2.68}{\meter} (using real fingerprints only), to min./max. positioning errors of \SI{0.04}{\meter}/\SI{2.23}{\meter} (using real and artificial fingerprints) & - & used \num{1} AP and \num{1} modified WNIC &  a classroom with \num{7}$\times$\num{7} training RPs of \SI{1}{\meter} spacing and \num{25} randomly placed testing RPs \\
\hline

\cite{wang2019robust} \num{2019} & improving \num{2}D coordinate regression when the time interval between collecting training and testing datasets is long & SDAE + FC; Keras & positioning errors of \SI{5.64}{\meter}, \SI{3.05}{\meter}, and \SI{4.24}{\meter} on the UJIIndoorLoc dataset, the authors' dataset (\num{0}-day time interval), and the authors' dataset (\num{52}-day time interval), respectively & CDFs are provided; positioning errors of $\leq$ \SI{8.0}{\meter} and \SI{6.0}{\meter} \SI{80}{\percent} of the time on the UJIIndoorLoc dataset and the authors' dataset (\num{52}-day time interval), respectively, and $\leq$ \SI{5.2}{\meter} \SI{90}{\percent} of the time on the authors' dataset (\num{0}-day time interval)& \SI{20.84}{\second} training time on a Central Processing Unit (CPU) and \SI{181}{\milli\second} response time & - & UJIIndoorLoc dataset, a \num{40}$\times$\SI{30}{\meter\squared} office space with \num{20} RPs of $\geq$ \SI{5}{\meter} spacing, and a \num{40}$\times$\SI{60}{\meter\squared} office space with \num{57} RPs of $\geq$ \SI{5}{\meter} spacing\\
\hline

\multicolumn{8}{c}{BLE Fingerprinting Solutions}\\
\hline
\cite{7959171} \num{2017} & \num{3}D BLE fingerprinting & DAE and $k$NN & horizontal/vertical mean error of \SI{1.089}{\meter}/\SI{0.341}{\meter} & horizontal/vertical CDFs are provided; horizontal/vertical errors of $\leq$ \SI{2.0}{\meter}/\SI{0.6}{\meter} \SI{92.0}{\percent} of the time; horizontal/vertical $\sigma$ of \SI{0.621}{\meter}/\SI{0.202}{\meter} & requires a measurement
window of \SI{1}{\second} before computing an output + \SI{0.84}{\milli\second} response time & deployed \num{10} BLE beacons & \SI{17.5}{\meter} by \SI{9.6}{\meter} conference room with \num{192} \num{3}D RPs at heights of \SI{0.8}{\meter}, \SI{1.6}{\meter}, and \SI{2.4}{\meter} \\
\hline
\cite{mohammadi2018semisupervised} \num{2017} & improving the localization accuracy by incorporating unlabeled BLE measurements & VAE and DRL; Keras for TensorFlow & \SI{4.3}{\meter} mean error & - & - & deployed \num{13} BLE beacons & \SI{60.96}{\meter} by \SI{54.86}{\meter} library floor \\
\hline
\cite{iqbal2018accurate} \num{2018} & tracking patients and clinical staff using a WSN and BLE tags & CNN and FC; Keras for TensorFlow & \SI{99.9}{\percent} classification accuracy; F\num{1}-score of \num{0.999} & precision of \SI{0.999} and recall of \SI{0.999} & \SI{30}{\minute} training time on a CPU & deployed \num{21} Raspberry Pi nodes and used \num{10} BLE wearable tags & clinical environment composed of \num{16} rooms and \num{5} hallways \\
\hline
\multicolumn{8}{c}{Cellular Fingerprinting Solutions}\\
\hline
\cite{8570849} \num{2019} & cellular RSS fingerprinting using data augmentation (continuous) & FC & median positioning error of \SI{0.78}{\meter} & positioning error of $\leq$ \SI{3}{\meter} \SI{90}{\percent} of the time ; CDF is provided & - & - &  \SI{11}{\meter} by \SI{12}{\meter} office space with \num{51} RPs \\
\hline
\cite{8446013} \num{2018} & massive MIMO \num{3}D fingerprinting based on channel coefficients (simulated and real data) & FC; TensorFlow & depending on several scenarios, various sub-meter positioning errors were reported & - & \si{2,136,067} parameters; \SI{12}{\hour} training time on a GPU & deployed a linear array of \num{16} antennas and used a special device to collect data &  \SI{20}{\meter} by \SI{7}{\meter} testbed \\
\hline
\cite{8292280} \num{2017} & massive MIMO fingerprinting based on channel structure (simulated data) & CNN & NRMSE of \num{0.6}$\lambda$ ($\lambda \approx$ \SI{1}{\meter}) & - & computational complexity is analytically derived & - & \num{25}$\lambda$ by \num{25}$\lambda$ testbed \\
\hline
\multicolumn{8}{c}{Magnetic Field and IMU Fingerprinting Solutions}\\
\hline
\cite{8115887} \num{2017} & a magnetic field landmark classifier & CNN & \SI{80.8}{\percent} classification accuracy & - & \SI{15}{\minute} training time on a GPU &  \SI{2}{\hour} for site surveying using a wheeled robot &  \SI{15}{\meter} by \SI{22}{\meter} testbed with \num{35} magnetic landmarks \\
\hline

\end{tabular}
\end{table*}

\begin{table*}[!t]
\tiny
\renewcommand{\arraystretch}{1.3}
\caption*{TABLE \ref{fingerprints_table_1} (continued)\\a summary and comparison of the reviewed fingerprinting solutions}
\label{fingerprints_table_2}
\centering
\begin{tabular}{|m{0.4cm}|m{1.6cm}|m{1.2cm}|m{2.5cm}|m{2.7cm}|m{2cm}|m{1.5cm}|m{2.7cm}|}
\hline
\bfseries Work and year & \bfseries Purpose & \bfseries Model and framework & \bfseries Accuracy & \bfseries Precision & \bfseries Complexity & \bfseries Cost & \bfseries Scalability\\
\hline

\multicolumn{8}{c}{Magnetic Field and IMU Fingerprinting Solutions (continued)}\\
\hline
\cite{8626558} \num{2018} & fingerprinting using single magnetic field fingerprints  & CNN; TensorFlow & \SI{97.77}{\percent} RP classification accuracy with \SI{13.6}{\cm} mean error & ECDF is provided; min./max. errors of \SI{0.0}{\meter}/\SI{40.76}{\meter}; $\sigma$ of \SI{1.7}{\meter} & \si{263,369} parameters; \SI{2.4}{\milli\second} response time on a CPU  & - & \SI{185}{\meter\squared} testbed (\num{3} rooms and \num{3} corridors) with \num{317} uniformly distributed and regularly spaced RPs \\
\hline
\cite{8560119} \num{2018} & a binary classifier for indoor corners & CNN and RNN & F\num{1}-score of \num{0.855} & precision of \SI{0.805} and recall of \SI{0.911} & requires a measurement window of \SI{2}{\second} before computing an output & - & collected data from two different smartphones \\
\hline
\cite{8254556} \num{2017} & fingerprinting using consecutive magnetic field fingerprints  & RNN; TensorFlow & \SI{1.062}{\meter} mean error & distribution of error is provided; min./max. errors of \SI{0.44}{\meter}/\SI{3.87}{\meter} & - & - & \SI{21.47}{\meter} by \SI{10.17}{\meter} testbed with \num{629} RPs\\
\hline
\cite{8660396} \num{2019} & a magnetic field landmark classifier & LSTM; TensorFlow & classification accuracies of \SI{91.1}{\percent} and \SI{97.2}{\percent}, and F\num{1}-scores of \num{0.90} and \num{0.97} for the corridor and lab, respectively & precisions of \SI{0.906 } and  \SI{0.97}, and recalls of \SI{0.911} and \SI{0.971} for the corridor and lab, respectively & - & - & two testbeds: a \si{100}$\times$\SI{2.5}{\meter\squared} corridor with \num{25} uniformly distributed landmarks in \num{1}D and a \num{7}$\times$\SI{7}{\meter\squared} lab with \num{17} uniformly distributed landmarks in \num{2}D\\
\hline

\multicolumn{8}{c}{Image Fingerprinting Solutions}\\
\hline
\cite{kendall2015posenet} \num{2015} & camera pose estimation from a single query image & CNN (based on GoogLeNet); Caffe & \SI{0.5}{\meter} median positioning error and \ang{5} median orientation error & CDF of positioning and orientation error is provided for two of the seven scenes & \SI{1}{\hour} training time and \SI{5}{\milli\second} response time on a GPU; it takes \si{50} MB to store the parameters & - & a multi-room testbed (\num{7}-Scenes dataset); scales to different cameras with unknown intrinsics; provides outdoor positioning but with lower accuracy \\
\hline
\cite{walch2017image} \num{2017} & enhanced camera pose estimation from a single query image & CNN (based on PoseNet) + LSTM; TensorFlow & \SI{0.31}{\meter} median positioning error and \ang{9.85} median orientation error on the \num{7}-Scenes dataset; \SI{1.31}{\meter} median positioning error and \ang{2.79} median orientation error on the authors' dataset & - & - & used a laser ranging system to provide ground truth information & a multi-room testbed (\num{7}-Scenes dataset); a university floor of \si{5,575} \si{\meter\squared}; provides outdoor positioning but with lower accuracy\\
\hline
\cite{7743683} \num{2016} & image-based symbolic positioning & CNN (based on AlexNet) + Naïve Bayes ; Caffe & \SI{95}{\percent} room classification accuracy & - & \SI{1.14}{\second} training time on a CPU (for Naïve Bayes only) & - &  a testbed with \num{16} rooms \\
\hline
\cite{ha2018image} \num{2018} & reduce site surveying efforts and provide indoor positioning using BIM images & CNN (based on VGG) + cosine similarity & \SI{91.61}{\percent} image retrieval accuracy & - & it takes \SI{1.14}{\milli\second} to perform cosine similarity between two feature vectors & - & a corridor with \num{14} arbitrarily placed RPs \\
\hline
\cite{ACHARYA2019245} \num{2019} & reduce site surveying efforts and provide camera pose estimation using BIM images & CNN (based on PoseNet); Caffe &  \SI{1.88}{\meter} median positioning error and \ang{7.73} median orientation error & - & it takes $\approx$ \SI{1}{\hour} for fine-tuning on a GPU; response times of \SI{5}{\milli\second} and \SI{0.625}{\second} on a GPU and CPU, respectively& - & a \SI{30}{\meter} long corridor \\
\hline

\multicolumn{8}{c}{Hybrid Fingerprinting Solutions}\\
\hline
\cite{7776928} \num{2017} & fusion of magnetic filed and image fingerprints for improved infrastructure-free positioning & CNN (Places-CNN) + FC + particle filtering & - & CDF is provided; positioning error of  $\leq$ \SI{1}{\meter} \SI{87}{\percent}, \SI{78}{\percent}, \SI{88}{\percent}, and \SI{89}{\percent} of the time, for lab, garage, canteen, and office, respectively & time complexity of $\mathcal{O}(n)$ for the online phase, where $n$ is the number of particles ($n$ was set to \num{2000}) & - & four indoor environments: a lab, a garage, a canteen, and an office with areas of \si{4,094}\si{\meter\squared}, \si{732}\si{\meter\squared}, \si{1,148}\si{\meter\squared}, and \si{2,193}\si{\meter\squared}, and regularly-spaced test RPs of \si{540}, \si{121}, \si{197}, and \si{215}, respectively; user heterogeneity\\
\hline
\cite{8354175} \num{2018} & improving camera pose regression by incorporating BLE fingerprints & dual-steam CNN; TensorFlow & \SI{0.60}{\meter} mean positioning error and \ang{4.8} mean orientation error & CDF is provided; position/orientation error of  $<$ \SI{1.1}{\meter}/\ang{7.1} \SI{90}{\percent} of the time; position/orientation $\sigma$ of \SI{0.6}{\meter}/\ang{7.8} & \SI{7}{\milli\second} response time on a GPU  & deployed \num{479} BLE beacons; used a LiDAR system to provide ground truth pose information & six indoor environments with areas of \si{2,646}\si{\meter\squared}, \si{1,280}\si{\meter\squared}, \si{3,480}\si{\meter\squared}, \si{812}\si{\meter\squared}, \si{1,353}\si{\meter\squared}, and \si{2,000}\si{\meter\squared} \\
\hline
\cite{8331081} \num{2018} & classification of locomotion activity for indoor positioning & SDAE & mean F\num{1}-score of \num{0.940} & F\num{1}-score can vary depending on a user's movement characteristics and the activity itself & requires a measurement window of \SI{2}{\second} before computing an output & - & briefly investigated user heterogeneity \\
\hline
\cite{8422562} \num{2018} & positioning using magnetic filed and visible light fingerprints for WiFi-deprived areas & LSTM; TensorFlow & - & CDF is provided; positioning error of $\leq$ \SI{2}{\meter} \SI{85}{\percent} and \SI{75}{\percent} of the time for the lab and corridor, respectively; max. error of \SI{3.7}{\meter} and \SI{6.5}{\meter} for the lab and corridor, respectively & - & - & two testbeds: a \si{6}$\times$\SI{12}{\meter\squared} lab with \num{12} RPs in \num{2}D and a \num{2.4}$\times$\SI{20}{\meter\squared} corridor with \num{10} RPs in \num{1}D \\
\hline
\cite{8554268} \num{2018} & alleviating local and global ambiguity by exploiting WiFi and magnetic field fingerprints& CNN; MatConvNet which is an implementation of CNNs for MATLAB\cite{vedaldi2015matconvnet} & - & CDF is provided; positioning error decreases with increased cell diameter, e.g, positioning error of $\leq$ \SI{2}{\meter} \SI{65}{\percent} and \SI{100}{\percent} of the time for diameters of \SI{1.3}{\meter} and \SI{26}{\meter}, respectively & \num{2}$\sim$\SI{3}{\second} to complete WiFi scans + \SI{2}{\milli\second} response time on a cloud computing platform & used the pre-existing infrastructure of \num{151} APs & a \si{60}$\times$\SI{40}{\meter\squared} office space; comparable positioning error with respect to \num{4} users of different heights\\
\hline
\cite{8612930} \num{2019} & combining Xbee, BLE, WiFi, and heading measurements for indoor tracking & FC & mean positioning error of \SI{0.45}{\meter} & - & - & deployed \num{5} Raspberry Pi nodes & a \si{31}$\times$\SI{21}{\meter\squared} testbed (\num{4} rooms and \num{2} corridors) with \num{70} RPs as rectangular cells of various sizes\\
\hline

\multicolumn{8}{c}{Miscellaneous Fingerprinting Solutions}\\
\hline
\cite{luo2016deep} \num{2016} & UWB fingerprinting using CIR parameters & DBN & - & positioning error of $<$ \SI{1.5}{\meter} \SI{90}{\percent} of the time; CDF is provided & - & simulated deploying \num{3} UWB receivers and an UWB emitter & simulated a \SI{12.5}{\meter} by \SI{22.4}{\meter} office environment \\
\hline
\cite{8693950} \num{2019} & visible light fingerprinting using RSS & FC & positioning errors of \SI{3.40}{\centi\meter}, \SI{4.35}{\centi\meter}, and \SI{4.58}{\centi\meter} for the diagonal, arbitrary, and even sets, respectively& - & computational
complexity is analytically derived; \SI{11.25}{\milli\second} training time and \SI{8.66}{\milli\second} response time using Intel XEON & deployed \num{4} LEDs, a modulator, and a photodiode & a \num{1.8}$\times$\SI{1.8}{\meter\squared} testbed with \num{100} uniformly distributed and equally spaced RPs, \num{20} of which were selected for training while all RPs were used for testing \\
\hline
\cite{8543833} \num{2018} & acoustic fingerprinting using spectrogram & CNN; TensorFlow & sector classification accuracy of \SI{98}{\percent} & - & \SI{2.1}{\hour} training time and \SI{9.3}{\second} response time & simulated deploying \num{4} microphones & simulated a \num{10.2}$\times$\SI{10.2}{\meter\squared} testbed of \num{9} equal sectors\\
\hline
\cite{8761800} \num{2019} & passive positioning using RFID readers and tags & DBN & RMSE of $\approx$ \SI{1}{\meter} & positioning error of $<$ \SI{2}{\meter} \SI{90}{\percent} of the time; CDF is provided &   & simulated deploying \num{6} RFID readers and \num{43} tags & simulated a \num{12}$\times$\SI{12}{\meter\squared} LoS testbed with \num{619} training RPs and \num{20} testing RPs\\
\hline

\end{tabular}
\end{table*}

\subsection*{DBN-based Solutions:}
Wang \textit{et al}. \cite{DeepFi1,DeepFi2} proposed \textit{DeepFi}, a system that employs CSI-amplitude fingerprints to provide indoor positioning using a single AP. The idea was to have a dedicated DBN for every training RP in the environment. Each DBN is trained only on the fingerprints collected at its corresponding location. Parameters are first initialized using the greedy algorithm and then fine-tuned using reconstruction loss. A single fingerprint represents a network packet sent by an AP and received by a modified WNIC. Both the AP and WNIC communicate via three antennas and a total of \num{90} values are extracted per packet, corresponding to \num{30} subcarriers per antenna. In the online phase, Bayes' Law obtains a posterior probability for every RP using a uniform prior and a likelihood based on a Radial Basis Function which takes, as input, the distance between the measured fingerprint and the reconstructed fingerprint. The final position estimation is taken as a weighted average of all RPs and their corresponding posterior probabilities. The experimental evaluation took place in a line-of-sight (LoS) environment (a \num{7}$\times$\SI{4}{\meter\squared} empty living room) and an NLoS environment (a \num{9}$\times$\SI{6}{\meter\squared} cluttered computer lab), where each environment is equipped with a single AP. A total of \num{50} RPs (\num{38} training and \num{12} testing) and \num{80} RPs (\num{50} training and \num{30} testing), arranged in a grid layout with \SI{0.5}{\meter} spacing, were used for the LoS and NLoS environments, respectively. Due to the harsher propagation conditions in the NLoS environment, better positioning accuracy was obtained for the LoS environment. Positioning errors of \SI{0.94}{\meter} and \SI{1.80}{\meter} were achieved for the LoS and NLoS, respectively. The positioning accuracy of DeepFi was compared with two benchmark schemes, \textit{FIFS} \cite{FIFS} and \textit{Horus} \cite{Horus}, which use probabilistic approaches for positioning based on CSI and RSS fingerprints, respectively. Overall, an increase in positioning accuracy of \SI{23}{\percent} and \SI{33}{\percent} was achieved over FIFS and Horus, respectively. However, the memory requirement of DeepFi grows linearly with additional training RPs since the weights of every DBN must be stored separately. Also, the heavy probabilistic implementation during the online phase resulted in a response time of \SI{2.56}{\second}, making DeepFi unfit for real-time positioning applications. The impact of different parameters on positioning accuracy was investigated; the authors concluded that denser training RPs, more antennas, and more packets, generally lead to improved accuracy.

 Wang \textit{et al}. also proposed \textit{PhaseFi} \cite{PhaseFi1,PhaseFi2}, a system that uses CSI-phase fingerprints to provide indoor positioning using a single AP. Unlike CSI-amplitude measurements, raw CSI-phase measurements cannot directly be used for positioning since they contain significant noise and phase offsets. Therefore, pre-processing is required to obtain calibrated phase measurements. PhaseFi was developed and tested under similar conditions to DeepFi. PhaseFi achieved a positioning error of \SI{1.08}{\meter} for the LoS environment and \SI{2.01}{\meter} for the NLoS environment, which is slightly inferior to that obtained by DeepFi. 

Later, Wang \textit{et al}. proposed \textit{BiLoc} \cite{BiLoc}, a system that combines CSI-amplitude fingerprints with the estimated Angle of Arrivals (AoAs) to enhance positioning accuracy. Both measurements were obtained using two modified WNICs operating in the \SI{5}{\giga\hertz} radio band. The authors resorted to this band because firmware limitations prevented them from obtaining accurate AoA measurements in the \SI{2.4}{\giga\hertz} band. The positioning scheme used in BiLoc is like that used in DeepFi and PhaseFi, except that BiLoc requires an additional DBN to process AoA information. A parameter, $\rho$, used in the online phase, controls the influence each modality has on the positioning. The evaluation settings are slightly different from those used in DeepFi and PhaseFi. The NLoS environment had fewer RPs (\num{15} training and \num{15} testing) since grid spacing increased to \SI{1.8}{\meter}. The LoS environment was replaced by a \num{2.4}$\times$\SI{24}{\meter\squared} corridor that had \num{20} RPs (\num{10} training and \num{10} testing) arranged in a straight line with \SI{1.8}{\meter} spacing. The impact of varying $\rho$ on positioning accuracy was investigated. For the NLoS environment, the authors concluded that both modalities should have equal influence since they complemented each other. However, for the LoS environment, given that all RPs were in \num{1}D, the influence of AoA should be minimized since AoA measurements have limited contribution in such a scenario. Overall, BiLoc outperformed DeepFi by \SI{24}{\percent} in positioning accuracy, but its prediction latency increased by \SI{70}{\percent}. Also, BiLoc requires double the amount of memory space compared to DeepFi or PhaseFi.

\subsection*{CNN-based Solutions:}
Chen \textit{et al}. \cite{8027020} proposed \textit{ConFi}, a system that uses CSI images and a single AP for indoor positioning. Since CNNs are best suited for image classification, the authors transformed CSI-amplitude measurements into RGB images and called them “CSI images.” The three channels of a CSI image correspond to the information obtained from three antennas. Each channel has \num{30}$\times$\num{30} pixels. Column pixels represent information extracted from the \num{30} subcarriers while row pixels represent the information extracted from \num{30} consecutive packets. To reduce the cost of site surveying, data augmentation was performed on the training set, increasing the positing accuracy by \SI{2.5}{\percent}. In the online phase, the positioning output is taken as a weighted centroid of the three highest-ranking RPs, as given by the softmax layer. The evaluation was performed in a \num{16.3}$\times$\SI{17.3}{\meter\squared} office space with five rooms, where a single AP was stationed in one of the rooms. \num{64} RPs, with spacing between \SI{1.5}{\meter} and \SI{2}{\meter}, and \num{32} randomly placed RPs were chosen for training and testing, respectively. A positioning error of \SI{1.36}{\meter} was reported, a \SI{9.2}{\percent} improvement over DeepFi, when evaluated in the same environment. 

The authors of DeepFi, PhaseFi, and BiLoc also developed \textit{CiFi} \cite{CiFi1,CiFi2}. Unlike these implementations, CiFi uses a single CNN for positioning. The network was trained on location-labeled AoA measurements that were transformed into images for CNN processing. A single image represents \num{60} measurements extracted from \num{60} packets (i.e., \num{60}$\times$\num{60} pixels). During the online phase, positioning is calculated as a weighted average of all RPs, using an input of \num{16} images. The evaluation settings are like that used for BiLoc. Compared to BiLoc, the positioning accuracy of CiFi is \SI{12}{\percent} lower, which is explained by the fact that CiFi utilizes only AoA measurements. The prediction latency of CiFi is comparable to BiLoc; however, its memory requirement is significantly lower. 

\subsection*{AE-based Solutions:}
Nowicki and Wietrzykowski \cite{nowicki2017low} used a Stacked AE (SAE), followed by an FC network, for multi-building and multi-floor classification. They indicated that previous approaches based on hierarchical processing \cite{7346967} have high complexity, requiring careful feature selection and a separate algorithm for each level of granularity (i.e., building then floor identification). The purpose of the SAE is to perform dimensionality reduction. This is important because a WiFi fingerprint has entries for all APs detected in an entire environment, but only a subset of these APs is observed for different locations. This is especially true for large-scale environments. The FC network maps the compact representation into its corresponding class, where a class represents a flattened label of a building-floor combination (e.g. ``\texttt{Building3-Floor5}''). The authors reported a \SI{92}{\percent} classification accuracy on the UJIIndoorLoc dataset.

Kim \textit{et al}. \cite{8215259,Kim2018} took it a step further by providing multi-building, multi-floor, and multi-location positioning. They argued that approaching this problem from a multi-class classification perspective, in which a separate class is created for every distinct location, is not scalable. Instead, they exploited the hierarchical nature of the problem by casting it as a multi-label classification problem.  Inspired by \cite{nowicki2017low}, they used an SAE for dimensionality reduction followed by an FC network for multi-label classification. A weighted loss function was used for the FC network, which penalized building misclassification more than floor misclassification and floor misclassification more than location misclassification. The building, floor, and location were predicted in parallel, entailing post-processing to remove invalid class combinations. The final location estimate was taken as a weighted centroid of valid combinations. Results were reported on the UJIIndoorLoc dataset. The dataset has \num{933} distinct locations, requiring \num{933} output nodes for multi-class classification. However, through their approach, the authors reduced this number to \num{118}. The reported results were \SI{99.82}{\percent} building hit rate, \SI{91.27}{\percent} floor hit rate, and \SI{9.29}{\meter} positioning error, results that did not prevail over state-of-the-art implementations \cite{7346967} (i.e., \SI{100}{\percent} building hit rate, \SI{93.74}{\percent} floor hit rate, and \SI{6.20}{\meter} positioning error).

To mitigate the effects of fluctuating RSS fingerprints, Zhang \textit{et al}. \cite{ZHANG2016279} proposed using a Stacked Denoising AE (SDAE). Their approach involved a two-step training strategy (i.e., unsupervised training followed by supervised training). The SDAE was first pre-trained on corrupted fingerprints (to emulate random noise) and then fine-tuned on labeled fingerprints. In the online phase, the output of the SDAE, given by the softmax layer, was fed to an HMM. The HMM enforced temporal coherence by considering the SDAE’s previous estimates. Indoor and outdoor testbeds were used for evaluation. The indoor testbed represents a building’s floor that has \num{91} RPs with \SI{1.8}{\meter} spacing, whereas the outdoor testbed represents a campus’s lawn that has \num{105} RPs with \SI{2}{\meter} spacing. The number of detected APs for the indoor and outdoor testbeds were \num{163} and \num{359}, respectively. It was expected that a better positioning accuracy would be obtained for the outdoor testbed, given the increased number of APs, however, comparable positioning errors were achieved in both testbeds (i.e., RMSE of $\approx$ \SI{0.37}{\meter}). This was ascribed to the increased distance between the smartphone and the APs since the outdoor testbed did not contain APs but rather relayed on the signals coming from nearby buildings. The authors demonstrated the superiority of their implementation over $k$NN and Support Vector Machine (SVM) and illustrated how the positioning accuracy of these classifiers tended to decrease as the training set increased. However, the authors didn’t discuss the implication of using HMM on positioning latency. Instead, they stated that the SDAE takes \SI{0.25}{\second} to produce a location estimate.

Unlike the previously discussed AE-based positioning solutions, Wang \textit{et al}. \cite{wang2019robust} treated indoor positioning as a regression problem. The authors used an SDAE to extract time-independent features which were then fed to an FC network for \num{2}D coordinate regression. The most notable characteristic of their model is its ability to produce accurate positioning estimates even when there is a large time interval between the collection of training and testing datasets. Evaluation was performed using three datasets of varying time intervals that ranged from \num{0} to \num{52} days. Experimental results demonstrated that the authors’ model achieves comparable performance with a tree-fusion-based regression model when the time interval is short. However, in long time intervals, the authors’ model reduced the positioning error by up to \SI{14.7}{\percent}. Positioning errors of \SI{5.64}{\meter}, \SI{3.05}{\meter}, and \SI{4.24}{\meter} were reported on the UJIIndoorLoc dataset, the authors' dataset with a \num{0}-day time interval, and the authors' dataset with a \num{52}-day time interval, respectively.

\subsection*{RNN-based Solutions:}
Hoang \textit{et al}. \cite{8830368} exploited consecutive RSS fingerprints to alleviate the problem of location ambiguity often associated with one-shot positioning. They experimented with different RNN configurations and found that a feedback configuration, in which the location prediction of the previous time step is used as an input for the current time step, yields the best results. However, this configuration requires that the initial ground truth location of the user be provided at the beginning of a trajectory. The authors also tried different RNN types (i.e., Long Short-Term Memory (LSTM), Gated Recurrent Unit (GRU), Bidirectional, and vanilla RNN). Although all types produced comparable results, the best was obtained by LSTM, followed by GRU, Bidirectional, and vanilla RNN, respectively. Training is performed using trajectories (sequences) of  \num{10} (RP,RSS) tuples. \si{20,000} random training trajectories were generated on a \si{16}$\times$\SI{21}{\meter\squared} university floor with four corridors. A wheeled robot, carrying a smartphone and equipped with LiDAR, was used for data collection and ground truth measurements. The robot sampled from \num{365} RPs for training and \num{175} RPs for testing. Pre-processing using the iterative-recursive-weighted-average filter \cite{jiang2015indoor} was performed. This increased the accuracy over using raw RSS measurements slightly. For online positioning, sliding window averaging was applied over the outputs at previous time steps.  Positioning errors of \SI{0.75}{\meter} and \SI{4.2}{\meter} were reported on the authors' and the UJIIndoorLoc datasets, respectively. On average, this is a \SI{50}{\percent} reduction in positioning error over one-shot positioning that uses an FC network. 

\subsection*{GAN-based Solutions:}
Li \textit{et al}. \cite{8891678,8644149} proposed the use of GANs to reduce the cost of site surveying. The idea was to collect a small number of fingerprints and use GANs to generate more fingerprints. At each training RP, \num{500} CSI-amplitude packets were collected, \num{100} of which were randomly selected \num{1000} times to create amplitude/subcarrier plots. The amplitudes of \num{30} subcarriers from three antennas were used for this purpose. These plots were fed to a GAN to generate an additional \num{1000} plots. Both the generator and discriminator were CNNs. This process was repeated until all training RPs had been covered (i.e., \num{7}$\times$\SI{7} RPs with \SI{1}{\meter} spacing inside a classroom). Twenty testing RPs were randomly placed to evaluate the influence expanding the training set had on positioning accuracy. Employing an SVM classifier, the results revealed an \SI{11.94}{\percent} improvement over the initial training set that consisted of only real fingerprints. One drawback is that a separate GAN must be trained for every RP. One suggestion is to use a Conditional GAN (CGAN) \cite{mirza2014conditional}. This could reduce complexity because only a single CGAN is trained for all RPs and fingerprints are generated by conditioning on a specified RP.

\subsection*{2) BLE Fingerprints}

\subsection*{AE-based Solutions:}
To deal with fluctuating BLE measurements, Xiao \textit{et al}. \cite{7959171} utilized Denoising AEs (DAEs). They deployed ten BLE beacons along the walls of a \SI{17.5}{\meter} $\times$ \SI{9.6}{\meter} room and collected BLE fingerprints in a \num{3}D grid layout. A total of \si{192} \num{3}D training RPs (\si{8}$\times$\si{8}$\times$\si{3}) were used. Starting at a height of \SI{0.8}{\meter}, each RP was separated by \SI{0.8}{\meter} from adjacent horizontal/vertical RPs. During the offline phase, a DAE for each RP was trained on corrupted BLE fingerprints. The distance between the measured fingerprint and the reconstructed fingerprint for all DAEs was then used during the online phase by a Weighted $k$NN (W$k$NN) algorithm to estimate the user's location. The downside of this approach is its high memory requirement since the parameters of each trained DAE must be stored separately. The authors demonstrated the effectiveness of DAEs over AEs for BLE fingerprinting which they attributed to the superior ability of DAEs in capturing the statistical dependencies among noisy BLE measurements. They reported horizontal and vertical mean errors of \SI{1.089}{\meter} and \SI{0.341}{\meter}, respectively, on \num{128} \num{3}D testing RPs (\si{8}$\times$\si{8}$\times$\si{3}) located at heights of \SI{1.2}{\meter} and \SI{1.9}{\meter}.

Providing large amounts of location-annotated samples for training is not always attainable. To address this issue, Mohammadi \textit{et al}. \cite{mohammadi2018semisupervised} proposed a general framework for enhancing the accuracy of fingerprinting systems by incorporating unlabeled samples. The framework adopts semi-supervised learning into deep reinforcement learning (DRL). The semi-supervised part of the framework consists of a Variational AE (VAE) that carries out the task of annotating the unlabeled samples. These samples, along with the originally labeled ones, are then used to optimize the parameters of a DRL model that performs a series of moves to reach a target location. To validate the framework, the authors used a small set of labeled data and a much larger set of unlabeled data. The data represents raw and pre-processed measurements from \num{13} BLE beacons deployed in a \si{3,344} \si{\meter\squared} area. The authors reported an improvement of \SI{23}{\percent} on the localization accuracy compared to a supervised DRL model that only used the set of labeled data. 

\subsection*{CNN-based Solutions:}
Iqbal \textit{et al}. \cite{iqbal2018accurate} investigated the feasibility of using a Wireless Sensor Network (WSN) to track patients and clinical staff inside a clinical environment. They deployed \num{21} Raspberry Pi nodes inside \num{21} zones where each zone was a specific room or hallway. These nodes captured the signals emanating from wearable BLE tags broadcasting at a rate of  \SI{10}{\hertz}. The received signals were annotated with their zones and transformed into grayscale images where each image represented \SI{1}{\second} of RSS measurements corresponding to a given tag. The images were used to train a CNN to perform zone classification. The authors reported a classification accuracy of \SI{93.7}{\percent}. By using a sliding window of \num{30} consecutive predictions (i.e., considering temporal information) to train a separate FC network, the accuracy rose to \SI{99.9}{\percent}.

\subsection*{3) Cellular Fingerprints}

\subsection*{FC-based Solutions:}
Rizk \textit{et al}. \cite{8570849} proposed \textit{CellinDeep}, a system that uses an FC network to perform cellular RSS fingerprinting. To improve position estimation, an additional averaging step is performed, taking into consideration the position estimates of five successive samples. Moreover, data augmentation techniques were used, increasing the training set by \num{8}-fold, resulting in a \SI{70}{\percent} reduction in positioning error. They achieved a positioning error of less than \SI{3}{\meter} \SI{90}{\percent} of the time which, according to the authors, is a significant improvement over stare-of-the-art cellular RSS fingerprinting systems employing the $k$NN and SVM algorithms. They reported that cellular modems consume around \SI{90}{\percent} less power than WiFi modules when used for positioning. The positioning accuracy obtained by using different cellular providers with different BS densities was investigated. They found that the accuracy is proportional to the BS density, the higher the density, the higher the accuracy. The impact of device heterogeneity on accuracy was also investigated. For example, using the samples collected from six different smartphones reduced the positioning error by \SI{60}{\percent} as opposed to using the samples collected from a single smartphone.

Massive Multiple-Input Multiple-Output (MIMO) is an enabling technology for \num{5}G cellular networks. The idea is to use large antenna arrays (typically \num{8}$\times$\num{8} antennas) at BSs to deliver unprecedented communication benefits \cite{BJORNSON20193}. Arnold \textit{et al}. \cite{8446013} used a linear array of \num{16} antennas installed in a \SI{20}{\meter} by \SI{7}{\meter} area for indoor positioning. They used an FC network to correlate the antennas' channel coefficients to a \num{3}D position relative to the array's location. To avoid the burden of collecting a large dataset for training, a two-step training procedure was followed. First, the network was pre-trained on simulated LoS channel coefficients; then, it was fine-tuned with a small number of real LoS and NLoS measurements collected using a special probe. Various sub-meter accuracies were reported based on the environment setting (LoS vs. NLoS), the number of samples for fine-tuning, and the samples' spatial locations. Recently, the authors proposed a novel channel sounder architecture that they used to generate position-tagged massive MIMO measurements \cite{8661318}.

\subsection*{CNN-based Solutions:}
Vieira \textit{et al}. \cite{8292280} used a CNN to learn the structure of massive MIMO channels for indoor positioning. A cellular channel model (the COST \num{2100} \cite{6393523}) was used to generate unique channel fingerprints for each training/testing position. These fingerprints represent clusters of multipath components obtained from a BS equipped with a linear array of antennas. The fingerprints were transformed into an angular-delay domain to resemble sparse \num{2}D images that were  then used in training a CNN to regress the receiver's \num{2}D coordinates. The authors reported distance in terms of wavelength ($\lambda$). The RMSE, normalized by $\lambda$, was used as an accuracy metric where the achieved RMSE was \num{0.6}$\lambda$ inside a \num{25}$\lambda\times$\num{25}$\lambda$ confined area. Since all measurements were based on simulated data, real-world measuring impairments such as noise, channel fading, and mislabeling were not considered. The authors demonstrated that their implementation outperformed a non-parametric approach based on a grid search in both computational complexity and positioning accuracy. However, when the spacing between training samples is less than \num{0.5}$\lambda$, the non-parametric approach yields better positioning accuracy.

\subsection{Magnetic Field and IMU Fingerprints}

\subsection*{CNN-based Solutions:}
Lee and Han \cite{8115887} tried to improve on the work of \cite{gozick2011magnetic} by extending the concept of magnetic landmarks to \num{2}D spaces instead of only corridors. Unlike \cite{gozick2011magnetic}, the locations of the environment's steel structures did not need to be known; alternatively, a location was considered to be a magnetic landmark if the magnetic field intensity of that location was either lower or higher than predefined thresholds. For each landmark, a sequence of magnetic field measurements was transformed into a recurrence plot suitable for CNN processing. The sequence's length (in meters) and its trend (e.g., monotonic increase or decrease) were taken as auxiliary features to aid in the landmark classification process. Their approach was based on inferring the user's location if a landmark was classified correctly. However, how close or far the user was from the landmark was not specified; instead, a landmark classification accuracy of \SI{80.8}{\percent} was reported. 

Alhomayani and Mahoor \cite{8626558} used a publicly available dataset \cite{barsocchi2016multisource} to build a fingerprinting system for smartwatches. They treated the indoor localization problem as a multi-class classification problem. The system took a single, raw magnetic field and orientation measurement as input and produced a specific location as an output (i.e., one-shot positioning). While the use of magnetic distortions as fingerprints was already established, the use of orientation information was not. The authors applied the Maximal Information Coefficient to establish that a relationship exists between raw orientation measurements and certain indoor locations. This was explained by observing human traffic patterns inside corridors and how they tend to follow a counterclockwise motion. The authors performed model selection and concluded that the best performing model consists of two convolutional layers, two FC layers, and a softmax layer. A location classification accuracy of \SI{97.77}{\percent} with a mean error of \SI{13.6}{\centi\meter} was reported. However, one-shot positioning has the downside of increasing the maximum positioning error produced by the system. The authors reported a maximum error of \SI{40.76}{\meter}.

\textit{Fingerprint crowdsourcing} has recently been adopted to relieve the burden of site surveying by delegating this process to common users traversing the area \cite{ParkandDorothy}. This, however, has created a new set of challenges; among them is annotating the casually collected fingerprints with their true locations. One approach to addressing this issue is identifying corners which will serve as landmarks to facilitate fingerprint annotation. To this end, Wang \cite{8560119} proposed \textit{CRNet}, a deep learning network that identifies corners from pedestrian trajectories. The idea was based on recognizing the changes that magnetometer and IMU signals undergo when turning a corner. The network consisted of a CNN followed by an RNN. The network was fed a two-second window of raw magnetometer and IMU measurements at \SI{50}{\hertz} and it had to decide whether the window corresponds to a corner or not (i.e., binary classification). The author reported an F\num{1}-score of \num{0.855} on a trajectory dataset containing fake corners (i.e., turnarounds and turns that do not correspond to physical corners).

\subsection*{RNN-based Solutions:}
Jang \textit{et al}. \cite{8254556} proposed using continuous magnetic field measurements to reduce the positioning ambiguity associated with one-shot positioning. The authors chose RNNs because of their ability to capture the spatial/temporal dependency of magnetic field measurements in a given sequence. They first built a magnetic map of the \SI{218}{\metre\squared} testbed with \num{629} RPs, then used it to generate \si{50,000} routes using the random waypoint model. Each route was composed of \num{20}-step movements where the  coordinates of each step were associated with the observed magnetic field. \SI{95}{\percent} of the routes were used to train while the remaining \SI{5}{\percent} were used for evaluation. The reported localization error ranged from \num{0.44} to \SI{3.87}{\meter} with an average error of \SI{1.06}{\meter}. This is a \SI{66.18}{\percent} improvement over a BLE fingerprinting system deployed in the same area. However, the drawback, when compared to one-shot positioning, is that a user must initially walk \num{20} steps before his/her position can be estimated.

Bhattarai \textit{et al}. \cite{8660396} used LSTMs as magnetic landmark classifiers. At each landmark, a continuous sample of magnetic field measurements was recorded in all directions and then segmented into subsequences of equal length for training and testing. The output of each time step was combined and passed to a softmax layer that output the landmarks' probabilities. Two testbeds were used for evaluation, a \num{100}$\times$\SI{2.5}{\meter\squared} corridor with a \num{1}D layout of \num{25} uniformly distributed landmarks and a \num{7}$\times$\SI{7}{\meter\squared} lab with a \num{2}D layout of \num{17} landmarks. The authors found that a vanilla LSTM yielded the best result for the \num{1}D setting (\SI{91.1}{\percent} accuracy with a sequence of \num{16} measurements) while a bidirectional LSTM yielded the best result for the \num{2}D setting (\SI{97.2}{\percent} accuracy with a sequence of four measurements). The better result obtained in the latter testbed was attributed to the many electronic devices contained in it, which further distorted the magnetic field. The effectiveness of the implementation was demonstrated over $k$NN, SVM, and Decision Tree classifiers. Nonetheless, what constitutes a magnetic landmark in the authors' view was not clear, since landmarks appeared to be uniformly distributed RPs.

\subsection{Image Fingerprints}

\subsection*{CNN-based Solutions:}
In \num{2015}, Kendall \textit{et al}. \cite{kendall2015posenet} proposed \textit{PoseNet}, the first implementation to consider deep learning for real-time camera pose regression. PoseNet is based on \textit{GoogLeNet} \cite{7298594}, a CNN that has achieved great success in image classification tasks. The authors modified GoogLeNet to output position and orientation information relative to an arbitrary global reference frame. Transfer learning was leveraged by pre-training GoogLeNet on the \textit{Places} dataset \cite{Zhou2014}, a large scene-recognition dataset. Pre-training enabled PoseNet to preserve pose information in the intermediate representations, resulting in faster convergence and lower error than when training from scratch. For fine-tuning, the authors utilized Structure-from-Motion (SfM) to label images with camera pose. Compared to localizing with SIFT, PoseNet occupied far less memory and provided significantly faster estimates. Positioning and orientation errors of \SI{0.5}{\meter} and \ang{5} were reported on the \num{7}-Scenes dataset, respectively.

Walch \textit{et al}. \cite{walch2017image} modified PoseNet to include LSTM units right before outputs were produced. These units performed structured dimensionality reduction, leading to improvements in pose estimation. Fine-tuning was performed using a specially created dataset of wide-angle images taken $\approx$ \SI{1}{\meter} apart inside a \si{5,575} \si{\meter\squared} university floor. These images were annotated with ground truth pose information using a laser ranging system. The authors used their dataset, in addition to the \num{7}-Scenes dataset, to compare the performance of their implementation to PoseNet and a state-of-the-art SIFT-based method \cite{7572201}. The SIFT-based method outperformed both CNN-based methods on the latter dataset but failed on the former. This is because SIFT-based methods require accurate \num{3}D models to produce pose estimates, something the authors were not able to render due to the textureless surfaces and repetitive structures of the indoor environment. Overall, the positioning error was reduced by $\approx$ \SI{30}{\percent}, on both datasets, compared to PoseNet.

Symbolic positioning has also been investigated using feature point detectors \cite{Kawaj2010,6071954}. Werner \textit{et al}. \cite{7743683} utilized the CNN-based \textit{AlexNet} \cite{krizhevsky2012imagenet} as a generic feature extractor to classify a query image to one of sixteen rooms. They called their system Deep Mobile Visual Indoor Positioning System (\textit{DeepMoVIPS}). No fine-tuning was performed on the pre-trained network; instead, the authors directly fed the features extracted by the first FC layer to a Naïve Bayes classifier. These features helped their model generalize local to global views (i.e., from small views in training to large views in testing) well. However, this didn't hold for the opposite case due to the spatial invariance of features introduced by the CNN. A room classification accuracy of \SI{95}{\percent} was reported using global views for both training and testing.  

Ha \textit{et al}. \cite{ha2018image} proposed using \textit{Building Information Modeling} (BIM) (i.e., digital representations of a building's physical and functional characteristics), to reduce the effort of site surveying. The idea was to use synthetic images rendered from a \num{3}D BIM model without having to physically survey the area. However, cross-domain image retrieval was needed since positioning was performed using real images. Since SIFT-based matching failed, the authors utilized a pre-trained \textit{VGG} network \cite{simonyan2014very} for feature extraction. Features extracted by the fourth pooling layer were matched to those of pre-stored BIM images through cosine similarity. The BIM image with features closest to the features of the query image was retrieved and, hence, the user's position was determined. An image retrieval accuracy of \SI{91.61}{\percent} was reported inside a corridor with \num{14} RPs. While this approach does not require training, the retrieval technique followed makes the response time proportional to the BIM dataset size.

Similarly, Acharya \textit{et al}. \cite{ACHARYA2019245} proposed \textit{BIM-PoseNet}, a PoseNet fine-tuned with BIM rendered images. A virtual camera, with the same intrinsic parameters as the real camera used for testing, was utilized to tag these images with pose information. Cross-domain positioning was realized by performing gradient edge-detection on both synthetic and real images before feeding them to the network. A median positioning and orientation error of \SI{1.88}{\meter} and \ang{7.73}, respectively, was reported inside a \SI{30}{\meter} long corridor. Pose errors were mainly caused by indoor objects that appeared in the real images (e.g., furniture, decorations, etc.) but were not included in the BIM model.

\subsection{Hybrid Fingerprints}

\subsection*{CNN-based Solutions:}
Liu \textit{et al}. \cite{7776928} proposed \textit{VMag}, a fingerprinting system that fuses magnetic field fingerprints with image fingerprints. The authors offered two reasons for choosing these fingerprint types. First, both types do not depend on infrastructure for operation. Second, each type complements the other since magnetic field fingerprints are more prone to global ambiguity while image fingerprints are more prone to local ambiguity. In their implementation, deep learning was only leveraged for feature extraction. Image features were extracted using a pre-trained \textit{Places-CNN} \cite{Zhou2014} and fed to a separate FC network that had the task of fusing them with magnetic field features. The fused features, along with the user’s step length and heading information (inferred from IMU measurements), were passed to a particle filter that output the user's location. Experiment results based on four different indoor environments with a combined area of \si{8,167} \si{\meter\squared} and \si{1,073} test RPs, showed a positioning error of  $\leq$ \SI{1.0}{\meter} \SI{78.0}{\percent} to \SI{89.0}{\percent} of the time, depending on the environment. Nevertheless, the use of particle filtering adversely affected positioning latency since convergence to a reasonable positioning accuracy took as long as \SI{20}{\second}.

Ishihara \textit{et al}. \cite{8354175} improved the accuracy of camera pose regression by incorporating BLE RSS readings. They proposed a dual-stream CNN where one network regresses poses from images while the other simultaneously regresses poses from BLE measurements. The estimations of both networks were fused using a single output layer. The weights of the former network were initialized using the Places dataset while the weights of the latter network were randomly initialized. A smartphone was used to collect the training data which consists of (image,BLE) tuples labeled with ground truth pose information using a LiDAR system. An evaluation was conducted in six different indoor locations, with a total of \num{479} BLE beacons. Compared to PoseNet, which only uses images for pose estimation, positioning accuracy was improved by \SI{40}{\percent} but orientation accuracy deteriorated by \SI{30}{\percent}. The authors didn’t offer any explanation as to why the orientation estimation degraded.

Shao \textit{et al}. \cite{8554268} exploited the fact that WiFi fingerprints are generally immune to global ambiguity but prone to local ambiguity, while the contrary is true for magnetic field fingerprints. They proposed an architecture that harnesses the advantages of both fingerprint types through a two-branch, two-step training strategy. The output of two different CNNs, one for each fingerprint type, was fed to a subsequent CNN that output the final position estimation. Compared to using a single network, where the fingerprint types are concatenated and fed as a single input, this strategy yielded better results at the expense of a prolonged training process. RPs were represented as circular cells where the fingerprints gathered inside a cell were labeled with the coordinates of the cell's center. Investigating the impact of cell size on positioning error revealed that a considerable gain over (WiFi alone, magnetic field alone) positioning was more noticeable for larger cells (i.e., with diameters of \SI{10}{\meter} and up). This suggests that this approach is more advantageous for large-scale environments. The authors briefly investigated user heterogeneity by experimenting with four participants of different heights and concluded that the difference in positioning error between participants was insignificant. A positioning error of $\leq$ \SI{2}{\meter} \SI{95}{\percent} of the time was reported inside a \si{2,400} \si{\meter\squared} area for a cell diameter of \SI{10}{\meter}.

\subsection*{RNN-based Solutions:}

Wang \textit{et al}. \cite{8422562} proposed \textit{DeepML}, a system that blends magnetic field fingerprints and visible light fingerprints to provide indoor positioning for places that have little or no WiFi coverage, such as underground parking lots. Visible light fingerprints were chosen because they share two important characteristics with magnetic field fingerprints (i.e., omnipresence and stability). For example, underground parking lots maintain a lighting infrastructure in which the generated light intensity does not change over time. DeepML uses a sliding window of magnetic field and light intensity measurements for input and maps it to the most likely location via an LSTM network. An additional step was introduced for online positioning, calculating a location as a weighted average of all locations in the environment and the system's previous estimations for each location. While such an approach could result in robust estimations, the authors did not discuss its implications on positioning latency. Also, the impact of noise on light intensity measurements caused by transient light sources, such as car headlights, was not investigated. The authors reported a positioning error of $\leq$ \SI{0.5}{\meter} \SI{60}{\percent} of the time for two testbeds; one with a \num{1}D layout of \num{10} RPs and the other with a \num{2}D layout of \num{12} RPs, where RPs in both testbeds were separated by \SI{1.6}{\meter}.

\subsection*{AE-based Solutions:}

Gu \textit{et. al}. \cite{8331081} approached indoor positioning from the perspective of \textit{locomotion activity recognition} (LAR). This stems from the idea that a user's symbolic location can be inferred by simply determining his/her activity. For example, if a user's activity is recognized as “\textit{using the elevator},” then the user must be in an elevator. Correctly identifying such activities can aid positioning systems in filtering out unlikely estimates. To this end, the authors utilized an SDAE that took a two-second window of accelerometer, gyroscope, magnetometer, and barometer measurements, at \SI{32}{\hertz}, and mapped it to \num{1} of \num{8} activities using a softmax layer. These activities included elevator up/down, stairs up/down, walking/stationary, running, and false motion (the user remains stationary while using the phone for texting, gaming, etc.). The authors demonstrated that some activities are more distinguishable than others. For example, stairs up/down was sometimes misclassified as walking and vice versa. The authors also demonstrated that the combination of four sensors yielded higher accuracy than combinations of two/three sensors or individual sensors. An overall F\num{1}-score of \num{0.94} was reported; however, the authors showed that the accuracy can vary between users depending on their movement characteristics. While such an approach may not qualify as a stand-alone positioning technique, it could be used by crowdsourcing methods in identifying landmarks for fingerprint annotation.

\subsection*{FC-based Solutions:}

Belmonte-Hernández \textit{et al}. \cite{8612930} proposed \textit{SWiBluX}, an indoor tracking system that employs three wireless technologies: WiFi, BLE, and Xbee. The system uses special anchor nodes to collect the signal emitting from a custom-built wearable device. To account for measurement variability due to signal absorption by a user’s body, a user’s heading angle is attached to each measurement. This resulted in better position estimations compared to only using signal fingerprints. An FC network, which the authors showed to outperform $k$NN, SVM, and other shallow learning algorithms, was chosen as a location classifier. To filter out spurious predictions, the network's output was further refined using Gaussian outlier detection followed by particle filtering. While these post-processing phases improved the network's estimates by \SI{25}{\percent}, the corresponding computational overhead in the offline phase was inadmissible. Moreover, commodity smartphones currently do not provide support for Xbee. Experiments conducted inside a \SI{651}{\meter\squared} testbed, in which RPs were represented as rectangular cells of various sizes, demonstrated a mean positioning accuracy of \SI{0.45}{\meter}.

\subsection{Miscellaneous Fingerprints}
\subsection*{1) UWB Fingerprints}
\subsection*{DBN-based Solutions:}
Luo and Gao \cite{luo2016deep} used a DBN to track UWB emitters indoors. They exploited the fact that the channel impulse response (CIR) of UWB signals vary based on the geometry of the indoor environment and the distance between the emitter and receiver. They used some parameters of the CIR, such as the number of multipath components and the power of each path, as location fingerprints. Since their system was tacking-based, these parameters were extracted from several UWB receivers deployed in the environment. The DBN correlated the extracted parameters to the position of the UWB emitter. During training, the network was pre-tuned on unlabeled parameters then fine-tuned on location-labeled parameters. The authors reported a positioning error of $<$ \SI{1.5}{\meter} \SI{90}{\percent} of the time using only three UWB receivers in a \SI{280}{\meter\squared} area. This result is better than the error achieved by an FC network of the same architecture. The authors attributed this to the extra pre-training step. However, it should be noted that all the experiments and data in their work are simulation-based.

\subsection*{2) Visible Light Fingerprints}
\subsection*{FC-based Solutions:}
Zhang \textit{et al}. \cite{8693950} used an FC network to regress the \num{2}D coordinates of a photodetector. Four LEDs served as signal sources; each with a unique modulation frequency (different from the background light frequency) to prevent interference between them. The RSS of each LED was extracted by Fourier transformation and input into the FC network. The evaluation was performed in a small \num{1.8}$\times$\SI{1.8}{\meter\squared} area in which the LEDs were mounted at a height of \SI{2.1}{\meter}. Out of the \num{100} uniformly distributed and equally spaced RPs, \num{20} were chosen for training while all \num{100} RPs were used for testing. The authors investigated the impact of training RP placement on positioning accuracy. Three settings were used: 1) arbitrary set: RPs were randomly placed in the area; 2) even set: RPs were uniformly distributed and equally spaced; 3) diagonal set: RPs were placed along two diagonals. The best result was attained by the diagonal set (\SI{3.40}{\centi\meter}), followed by the arbitrary set (\SI{4.35}{\centi\meter}), and the even set (\SI{4.58}{\centi\meter}). Compared to a lateration-based algorithm implemented in the same area, the positioning error achieved by the diagonal set is \SI{91}{\percent} lower.

\subsection*{3) RFID Fingerprints}
\subsection*{DBN-based Solutions:}
Jiang \textit{et. al}. \cite{8761800} utilized RFID readers and reference tags for passive positioning. In such an implementation, the users, carrying the tags, have no means of triggering positioning requests. Instead, the system tracks users’ positions through the readers. DBNs were employed for this purpose and training was performed as described by Wang \textit{et al}. in \cite{DeepFi1}. However, simulated RSS measurements generated by a logarithmic path loss model were used. The authors demonstrated that DBNs not only outperformed $k$NNs and W$k$NNs but are also more resistant to RSS noise, which they attributed to the robust feature learning ability of DBNs. The testbed consisted of a simulated \num{12}$\times$\SI{12}{\meter\squared} room with six readers deployed along two sides of the room (i.e., three readers on each side). A positioning accuracy $<$ \SI{2}{\meter} \SI{90}{\percent} of the time was achieved using \num{619} uniformly distributed training RPs and \num{20} randomly placed testing RPs. 

\subsection*{4) Acoustic Fingerprints}
\subsection*{CNN-based Solutions:}
Zhang \textit{et al}. \cite{8543833} used a CNN to perform sound source localization. They simulated a \num{10.2}$\times$\SI{10.2}{\meter\squared} area with four microphones placed at the corners. The idea was to convert the sound waveform captured by the microphones into spectrograms and then combine them into a single spectrogram (a \num{2}D image) that highlighted the amplitude and time-delay differences between the microphones for a given location. The area was divided into nine sectors where \si{1,100} images were generated per sector. The model generating the images took the reverberation effect and white noise into account. A sector classification accuracy of \SI{98}{\percent} was reported. The authors demonstrated the supremacy of their implementation, in terms of accuracy and prediction latency, over $k$NN and SVM. However, the network was designed for a single sound source, a specific smartphone ringtone. It would be interesting to see whether the network would generalize to other ringtones or even a human voice. 

\subsection*{Discussion}
This subsection summarizes the main characteristics of the different deep learning models employed by the surveyed solutions and discusses which model works best for which contexts.

AEs are a family of hourglass-shaped neural networks\cite{hinton2006reducing}. Thus, they have the same number of neurons in the input layer as the output layer. A typical AE is trained to reconstruct an input using an encoder-decoder approach that forces the network to encode (compress) the input into a latent code from which the input can be decoded (reconstructed). The latent code given by the bottleneck layer can be used for dimensionality reduction. Thus, many of the surveyed works have exploited this feature to address sparsity in WiFi and BLE fingerprints. Additionally, successful applications in mitigating the effects of fluctuating and noisy signal measurements were attained using SDAEs (common variants of AEs that are trained to reconstruct an input from a corrupted version of it \cite{vincent2010stacked}).

CNNs were first designed to combat the problem of shift, scale, and distortion variance when classifying high dimensional patterns \cite{lecun1998gradient}. Thus, they are powerful tools for extracting generic features form images. This is reflected by the fact most of the surveyed work on image-based indoor positioning employed CNNs as opposed to other deep learning models. Additionally, CNNs have proven successful in positioning applications where the input data are not necessarily images but rather transformed to resemble images.

DBNs are neural networks that use unsupervised learning to facilitate supervised learning \cite{hinton2006fast}. This is achieved through a two-step training process, i.e., unsupervised pre-training using unlabeled data followed by supervised fine-tuning using labeled data. Since location-annotated samples are more expensive to obtain than unannotated ones, DBNs are most suitable for positioning applications where only a small amount of annotated data exist.

GANs are emerging deep learning architectures that were first introduced in \num{2014} \cite{goodfellow2014generative}. They are used to generate high-quality synthetic data from existing authentic data. Thus, they were exploited in several of the surveyed works to reduce the cost of site surveying by expanding the training set with artificial fingerprints.

RNNs are popular deep learning models for dealing with sequential and time-series data \cite{lstmhssj}. The output of the network is a function of the current input at time $t$ in addition to the previous inputs at time $\tilde{t}<t$. Thus, they are most suitable for positioning applications where the position history of a user is incorporated to refine positioning estimates.

\section{Pitfalls and Challenges}
\label{sec6}
\subsection*{Pitfalls:}
Having reviewed the different solutions presented in the literature, this section details the common pitfalls that should be avoided when designing deep learning-based indoor fingerprinting systems.
\subsubsection*{\bf Ill-defined materials and methods}
\begin{figure}[!t]
\centering
\includegraphics[width=0.65\columnwidth]{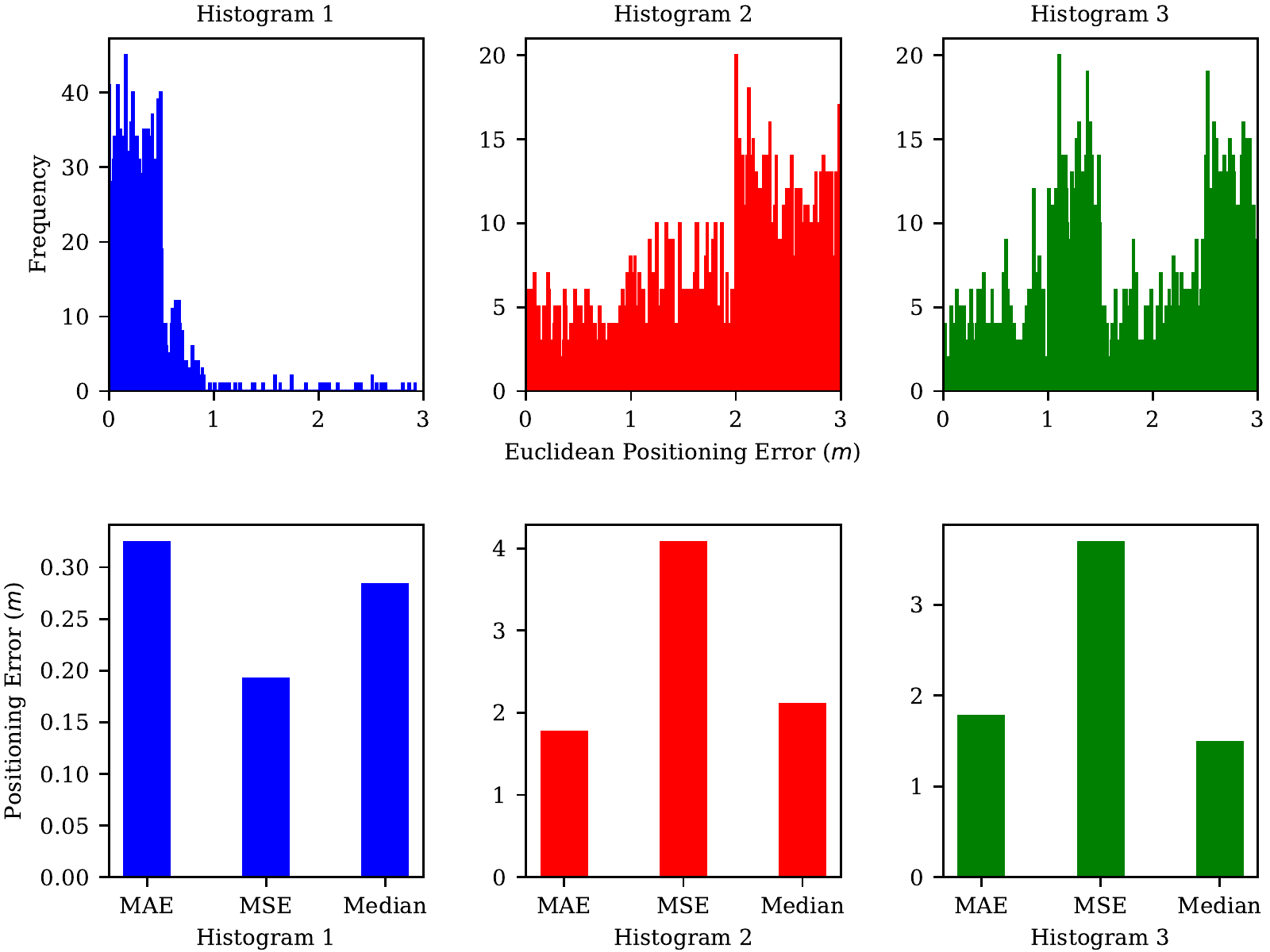}
\caption{The positioning error histograms of three different fingerprinting systems, calculated as Euclidean distance, and the corresponding MAE, MSE, and Median error. Each histogram was obtained using \si{1,850} testing fingerprints. All systems produce a min. error of \SI{0}{\meter} and a max. error of \SI{3}{\meter}.}
\label{MAE_MSE_Median}
\end{figure}

Full disclosure of an experiment is vital for the explanation, comparison, and reproducibility of indoor positioning research \cite{7346749}. This includes providing a detailed description of the data collection process and the testbed, the specifications of the proposed model, and the evaluation metrics used. Unfortunately, many of the reviewed solutions lack enough detail. For example, some works only use a single quantity to describe the performance of the system (i.e., positioning accuracy). Some go as far as to not specify how this quantity was calculated. Fig. \ref{MAE_MSE_Median} displays the positioning error histograms of three different systems. It seems that employing MSE to report accuracy was more tempting for the first system. Similarly, MAE and median error yielded the lowest error for the second and third systems, respectively. Therefore, without reporting the accuracy metric and the error distribution, it is difficult to fully characterize the performance of a given system. Moreover, other statistics such as minimum and maximum errors, when provided with a description of the testbed (e.g., size and location of RPs), could reveal a system's robustness to local/global ambiguity. Another pitfall is selective reporting. For instance, when reporting a system's response time, some works exclude the time it takes to sample the fingerprint, or the time needed for pre/post-processing and only report the prediction latency of the deep learning model.

\subsubsection*{\bf Collecting training and testing fingerprints from the same RPs}
The problem with having RPs serve a dual purpose is that testing may easily degenerate to benchmarking the system against its training samples \cite{loffler2018evaluation}. This is a common pitfall that could produce misleading results since over-fitting is much harder to detect in such cases. To avoid this problem and, hence, reflect the true generalizability of an implementation, the locations of testing RPs should be different from those of training RPs. For example, testing RPs could lay in between, but not on, the training RPs. 

\subsubsection*{\bf Post-processing that increases response time}
It is observed that whenever post-processing is involved, the real-time inference advantage offered by deep learning is often spoiled. It could be argued that the generally minor positioning accuracy gained by post-processing is not worth sacrificing positioning latency for. Therefore, one possible solution is to invest more effort into designing and fine-tuning the deep learning model instead of relying on quick, off-the-shelf models and let post-processing do the heavy lifting. This suggestion is backed by the fact that a lot of the reviewed solutions have achieved remarkable accuracies utilizing architecture design and hyperparameter tuning without resorting to post-processing.

\subsubsection*{\bf Evaluation under lab conditions}
Designing and evaluating a system under constrained/controlled settings (e.g., without the presence of moving people or constraining movements to predefined orientations/heights) limits its applicability. Also, such restrictions rarely apply in real-world use cases. Therefore, it is likely that the system's performance will deteriorate when tested under non-lab conditions since the fingerprints will be influenced by artifacts that were never part of the training data.

\subsubsection*{\bf Data leakage and sampling bias}
It has been noticed that some of the reviewed implementations unintentionally perform data leakage and sampling bias. The problem with these pitfalls is that they can result in practically incompetent and overly optimistic models. A common form of data leakage is normalizing a dataset before splitting it into training and testing sets while a common form of sampling bias is using an imbalanced dataset without employing the right performance metrics. Some useful resources on how to detect and avoid these subtle traps can be found in \cite{abu2012learning,Kaufman2012,lever2016points}.

\subsubsection*{\bf Not considering user/device heterogeneity} Different smartphones have different software/hardware components and different users have different carrying preferences, walking patterns, etc. Such variations cause measurements to be inconsistent, even when collected at the same RPs, leading to large positioning errors. While a few of the reviewed works hinted at this issue, no practical solution was offered to combat it. Resolving this issue is crucial because an ideal system would exhibit the same performance regardless of heterogeneity. Though it remains an active area of research, current proposals to address heterogeneity can be found in \cite{LaoudiasandRobert,7874080}. It is worth mentioning that some of these solutions are based on classical machine learning algorithms such as linear regression and expectation maximization; thus, it would be worthwhile to investigate applying deep learning to the heterogeneity problem.

\subsection*{Challenges:}
This section identifies the main challenges facing deep learning-based indoor fingerprinting systems. Solving these challenges is essential so that deep learning-based indoor fingerprinting can realize its full potential.
\subsubsection*{\bf Need for large datasets} Since deep learning involves tuning a considerable number of parameters, the need for large datasets is inevitable. Many of the reviewed systems demonstrated how predictive performance increases with increased training data. However, site surveying is notoriously laborious and costly. A recent strategy created to tackle this problem is crowdsourcing. Through crowdsourcing, the burden of fingerprint collection can be shared with users willing to participate either actively or passively \cite{ParkandDorothy}. However, crowdsourcing is not fully developed and much research is needed to address questions such as 1) What measures are needed to ensure the integrity of crowdsourced fingerprints? 2) What incentives are offered to users to engage and motivate them to participate in data collection? 3) How should inconsistent measurements caused by user/device heterogeneity be dealt with? and 4) How should fingerprints collected through passive crowdsourcing be annotated? Nevertheless, advances in crowdsourcing techniques will greatly facilitate the collection of data that is desperately needed to train deep learning models.

\subsubsection*{\bf Vulnerability to adversarial attacks} Many studies have demonstrated how deep learning models are vulnerable to adversarial attacks. Adversarial attacks refer to input samples designed by an attacker to force a trained model into producing incorrect predictions. Note that minute perturbations to the input features are enough to divert the network from functioning properly. This makes it harder to distinguish genuine inputs from adversarial ones. While the consequences of adversarial attacks are less severe in indoor positioning than in safety-critical applications (e.g., self-driving cars), neglecting this issue will have negative consequences. For example, an indoor positioning system that navigates a user to the wrong destination will likely lose the user’s trust. Recent attempts to tackle this problem include the work of Pei \textit{et al}. \cite{Pei3132785} who developed a framework called \textit{DeepXplore} to systematically generate adversarial inputs that can be used in retraining the model to fix its erroneous behaviors. Another example is the work of Yuan \textit{et al}. \cite{8611298} who identified \num{15} methods to generate adversarial inputs and described seven methods to countermeasure them.

\subsubsection*{\bf On-device deep learning} Most of the reviewed work used a smartphone for data collection; however, the smartphone was never used for data processing (i.e., calculation of the user's position is not carried out on the smartphone itself but rather offloaded externally). This is because deep learning models require sophisticated resources to run efficiently and research for devising ways to enable deep learning on smartphones is still ongoing. Recent developments in this area focus on designing Artificial Intelligence (AI) chips for smartphones. Examples of such chips include Apple’s \textit{Bionic Chip}, Qualcomm's \textit{Snapdragon 855}, Huawei’s \textit{Kirin 980}, and Samsung’s \textit{Exynos 9820}. In terms of software research, active areas include model compression \cite{han2015deep}, which aims at reducing complexity while preserving accuracy, as well as designing mobile-friendly deep learning architectures, such as \textit{MobileNet} \cite{8578572}, \textit{ShuffleNet} \cite{ma2018shufflenet}, and \textit{MnasNet} \cite{Tan_2019_CVPR}. From an indoor positioning perspective, enabling a user to use their own device to compute his/her position is desirable for four reasons. First, it increases the user’s privacy/security since no information must leave the user’s device. Second, it reduces latency since no communication is required with the server-side. Third, it increases availability since computation is performed in a decentralized fashion. Fourth, it reduces cost since a localization server is no longer needed. However, the impact that this approach has on a device's energy consumption needs to be investigated. 

\subsubsection*{\bf Interpretability and prediction uncertainty} A common challenge is the interpretability of deep learning models. Deep learning models are often treated as black-box function approximators, mapping one vector space into another. Thus, they do not provide any domain-specific insight or knowledge. This led practitioners in risk-averse domains to favor interpretable statistical models over accurate deep learning models. Deep learning also lags behind Bayesian modeling in terms of quantifying prediction uncertainty. In other words, deep learning models are incapable of obtaining principled uncertainty estimates. The root cause behind these issues is the lack of a coherent theoretical foundation mainly because research in deep learning has always revolved around improving state-of-the-art accuracy. In indoor positioning, interpretability and uncertainty quantification will ultimately be key factors in encouraging the adoption of deep learning in commercial settings. Luckily, researchers see a growing need to address these limitations and work in this area is continuing. Recent methods on improving interpretability and quantifying uncertainty are described in \cite{shrikumar2017learning,lundberg2017unified} and \cite{gal2016dropout,lakshminarayanan2017simple}, respectively. The practical value of incorporating such methods into future deep learning-based fingerprinting systems has yet to be appreciated.
\subsubsection*{\bf Privacy and security} Given the limited processing power of smart devices, the position estimation of deep learning-based fingerprinting systems is often performed on the server side. As a result, systems are capable of tracking users' movements and gathering their personal location information. Misuse of such information compromises users' privacy. Therefore, a system must clearly indicate what information is gathered from users and how this information will be used. Also, a system must indicate how users' information is protected against unauthorized access. Unfortunately, it was observed that privacy and security are not prioritized in the description of solutions and are generally overlooked. None of the reviewed solutions laid out how the users' location information is handled and protected. Privacy has become particularly concerning with the emergence of \textit{trajectory data mining} \cite{Zheng2015TDM}, which makes exploiting users' location information for advertising and other purposes inevitable. Positioning under the IoT brings another dimension to this problem since a user's location can be correlated with the location of IoT devices in the environment. This can also be exploited to reveal further information about the user's health, mood, and the activities he/she performs \cite{8692423}. Moreover, deep learning models themselves pose a threat to users' information because trained models can be attacked to identify the data records used in training \cite{7958568}. Unfortunately, there has been little research done on privacy issues in indoor localization. A recent proposal is the \textit{P\textsuperscript{3}-LOC} framework \cite{8542955}, a privacy-preserving, paradigm-driven framework for indoor localization. Other remedies include training deep learning models under differential privacy \cite{Abadi2016DLD} and reporting statistics from end-user client software, anonymously, with strong privacy guarantees \cite{Erlingsson2014RRA}. Nevertheless, with the ever-growing cyber security threats, privacy is becoming a major challenge that researchers should invest more effort in addressing.

\section{Conclusion and Future Perspectives}
\label{sec7}
The topic of indoor positioning has gained significant research interest in recent years due to the wide range of potential applications that it enables. Hence, it can easily be envisioned that indoor positioning will become a critical infrastructure in the foreseeable future. Amongst the different approaches to indoor positioning, the fingerprinting approach is the most investigated due to its simplicity, low cost, and high accuracy. Recent fingerprinting solutions leverage deep learning because it is a more powerful tool than traditional shallow learning. The objective of this paper was to provide a detailed review of these solutions. To this end, the reviewed solutions were classified based on the fingerprint type and further sub classified based on the deep learning model. All solutions belonging to a certain fingerprint type were compared using a well-defined performance evaluation framework. A detailed review of publicly available fingerprinting datasets suitable for training deep learning models was also included. Finally, the main pitfalls and challenges surrounding deep learning-based fingerprinting were discussed.  

Until the rise of a universally acknowledged solution for indoor positioning, the strive for better performance will continue to be a driving force behind the proposal of new solutions. The progress that has been made and is continuing to be made in sensing technologies, computing capabilities and deep learning will facilitate the development of these solutions. Also, since deep learning is being driven by not only academia but also industry, the development of new solutions will be further facilitated. For instance, a community of leading companies has recently introduced the Open Neural Network Exchange (ONNX) format \cite{ONNX}, an ecosystem that interchangeably represents deep learning models. In other words, ONNX enables models to be trained in one framework and transferred to another for inference. 

\subsection*{Future Perspectives:}
This section suggests future research directions for deep learning-based fingerprinting.
\begin{itemize}
\item It is easily deduced that publicly available datasets are major enablers of indoor positioning research. While very useful, current datasets have several limitations. First, they are limited to certain types of buildings (e.g., universities or research facilities). Few datasets directed at large-scale indoor structures such as airports, shopping malls, and convention centers, where indoor positioning is expected to be more profitable, exist \cite{8852722,lprgsg2020}. Second, they do not account for different transit modes (e.g., stairs, escalators, or elevators) or carrying modes (e.g., in-pocket, texting, or making a call). Third, they do not consider device heterogeneity because most of them utilized Android-based devices for data collection. It should be noted that Apple's iOS accounted for \SI{48.3}{\percent} of the smartphone market share in the U.S. (as of August \num{2019})\cite{statista}. Therefore, the proposal of datasets that can address these limitations will always be of great interest. 
\item Several emerging IoT wireless technologies will attract growing interest as localization enablers over the coming years. These include \textit{Sigfox}, \textit{LoRA}, \textit{NB-IoT}, and \textit{Wi-Fi HaLow}. Understanding their potentials and limitations for indoor localization will open new opportunities; yet, exploiting their characteristics for indoor localization largely remains unexplored. Currently, very few publications regarding these technologies and indoor fingerprinting \cite{8187642} can be found, whereas some initial studies are available for outdoor fingerprinting \cite{8533826,ijgi7110440}. Note that these works utilize classical machine learning algorithms; thus, it would be interesting to see how deep learning can push performance boundaries. 
\item While deep learning has recently seen implementations with indoor localization techniques such as pedestrian dead rocking \cite{8488496}, angulation \cite{8647687}, lateration \cite{8533766}, and device-free localization \cite{7765094}, its application to proximity detection and cooperative localization has yet to come.
\item Most of the image-based localization reviewed work used the same loss function originally proposed in PoseNet \cite{ kendall2015posenet}. Therefore, one possible research direction is to modify, or even propose, new loss functions for localization that can result in improved accuracy and/or lowered training time \cite{Kendall_2017_CVPR}.
\item Fingerprinting based on DRL is another area that has not received much attention. In fact, out of the reviewed work, only one solution utilized DRL. This is probably because DRL is a relatively new learning approach. However, DRL is gaining tremendous momentum and its application in a variety of fields, including communications and networking \cite{8714026}, is promising. In terms of deep learning architectures, some of the unexplored architectures for indoor positioning include, Residual Neural Networks \cite{he2016deep}, Convolutional  Autoencoders\cite{masci2011stacked}, and Conditional  Generative  Adversarial Networks\cite{mirza2014conditional}. 
\item The upcoming commercial deployment of \num{5}G promises ultra-reliable and low-latency communications (URLLC) \cite{8226757}. Furthermore, the reviewed solutions based on massive MIMO transmission demonstrated the potential for delivering centimeter-level positioning accuracy. However, massive MIMO-based solutions are proof-of-concept because they are based on simulations or hardware prototypes. It is expected that fingerprinting based on \num{5}G networks is going to be a new frontier that encompasses its own set of challenges and opportunities. Thus, one research direction is to investigate the feasibility of using real massive MIMO data emitting from real \num{5}G cellular networks for indoor positioning.
\end{itemize}

We hope that the information provided in this review assists investigators in better understanding deep learning-based fingerprinting systems, encourages new research efforts into this promising field, and paves the way for the practical deployment of systems as commercial products.

\bibliographystyle{IEEEtran}
\bibliography{Bibliography}

\end{document}